\documentclass[11pt,a4paper]{google}
\usepackage[authoryear,sort&compress,round]{natbib}
\usepackage{multirow,array,adjustbox}
\usepackage{pict2e}
\usepackage[section]{placeins}
\usepackage{amsmath,amsfonts,bm}

\def\eqref#1{equation~\ref{#1}}
\def\1{\bm{1}}

\DeclareMathAlphabet{\mathsfit}{\encodingdefault}{\sfdefault}{m}{sl}
\SetMathAlphabet{\mathsfit}{bold}{\encodingdefault}{\sfdefault}{bx}{n}

\newcommand{\E}{\mathbb{E}}

\definecolor{linkblue}{RGB}{0,0,139}
\definecolor{royalblue}{RGB}{65,105,225}
\usepackage[colorlinks=true,linkcolor=linkblue,urlcolor=royalblue,citecolor=brown,anchorcolor=blue]{hyperref}
\newcommand{\CorrespondingAuthorEmail}{kema@hkust-gz.edu.cn}
\DeclareRobustCommand{\CorrespondingAuthorSymbol}{%
  \href{mailto:\CorrespondingAuthorEmail}{\textcolor{black}{%
    \begingroup
    \setlength{\unitlength}{1em}%
    \linethickness{0.045em}%
    \begin{picture}(1.23,0.88)(-0.04,-0.04)%
      \polygon(0,0)(1.15,0)(1.15,0.8)(0,0.8)%
      \polyline(0,0.8)(0.575,0.23)(1.15,0.8)%
      \polyline(0,0)(0.38,0.423)%
      \polyline(0.77,0.423)(1.15,0)%
    \end{picture}%
    \endgroup
  }}%
}

\usepackage[capitalise,noabbrev]{cleveref}
\crefname{appendix}{appendix}{appendices}
\Crefname{appendix}{Appendix}{Appendices}
\renewcommand{\absfont}{\normalfont\linespread{1.2}\fontsize{11}{12}\selectfont}
\newcommand{\std}[1]{{$\pm$#1}}

\makeatletter
\renewcommand{\AB@affilsepx}{\protect\\[3pt]\protect\Affilfont}

\newcommand{\appendixtableofcontents}{%
  \begingroup
  \renewcommand{\contentsname}{Appendix Contents}%
  \@starttoc{atoc}%
  \endgroup
}

\newcommand{\startappendixtoc}{%
  \let\oldaddcontentsline\addcontentsline
  \renewcommand{\addcontentsline}[3]{%
    \def\@tempa{##1}\def\@tempb{toc}%
    \ifx\@tempa\@tempb
      \oldaddcontentsline{atoc}{##2}{##3}%
    \else
      \oldaddcontentsline{##1}{##2}{##3}%
    \fi
  }%
}
\makeatother

\title{More Camera Frames, Better Driving? Within-Scene Information Scaling \mbox{in Autonomous Driving}}
\author[1,*]{Yumao Liu}
\author[2,*]{Rui Qian}
\author[2]{Qiang Huang}
\author[2]{Jiahao Wei}
\author[1]{Tao Liu}
\author[1]{Xiangyu Li}
\author[1\,\CorrespondingAuthorSymbol]{Ke Ma}
\renewcommand{\Affilfont}{\normalfont\fontsize{11}{13}\selectfont}
\affil[ ]{{\normalfont\fontsize{10.5}{13}\selectfont
  \{\href{mailto:yumaoliu.ai@gmail.com}{\textcolor{black}{yumaoliu.ai}},
  \href{mailto:qiianruii@gmail.com}{\textcolor{black}{qiianruii\}@gmail.com}}}}
\makeatletter
\let\savedaffilseparator\AB@affilsepx
\renewcommand{\AB@affilsepx}{,\space\protect\Affilfont}
\affil[1]{\mbox{The Hong Kong University of Science and Technology (Guangzhou)}}
\let\AB@affilsepx\savedaffilseparator
\makeatother
\affil[2]{\mbox{Fudan University}}
\affil[*]{\fontsize{10.5}{13}\selectfont\,\textit{Equal contribution}\qquad
  \textsuperscript{\CorrespondingAuthorSymbol}\,\textit{Corresponding author}}
\correspondingauthor{Ke Ma (\href{mailto:\CorrespondingAuthorEmail}{\textnormal{\CorrespondingAuthorEmail}})}

\let\manuscriptabscontent\abscontent
\renewcommand{\abscontent}{%
  \noindent
  \begin{adjustbox}{max width=\linewidth,raise={2.13bp}{\height}{\depth}}%
    \normalfont\fontsize{12}{13}\selectfont\urlstyle{same}%
    \raisebox{-0.06em}{\includegraphics[height=1em]{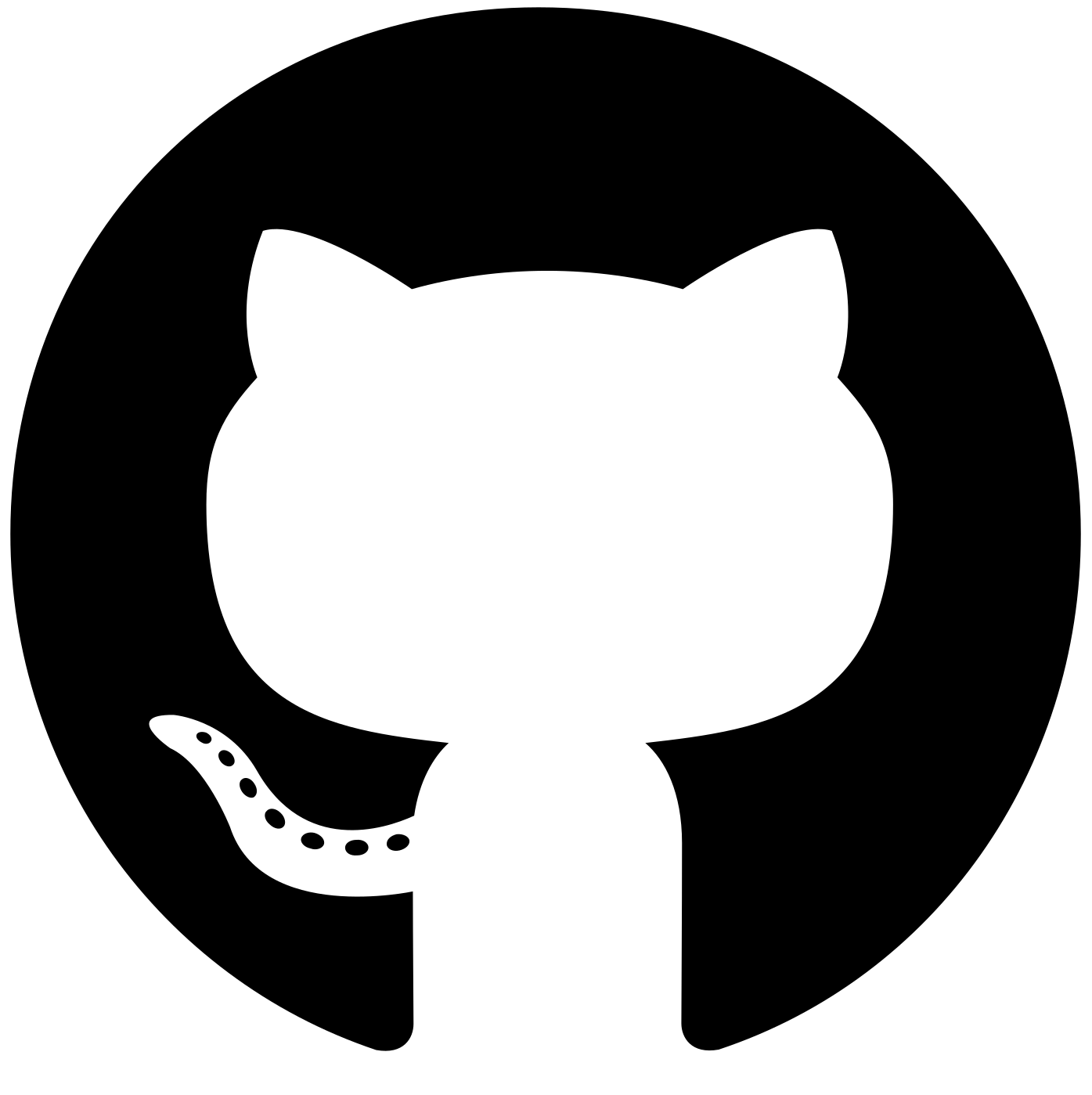}}\ Code:
    \url{https://github.com/KITE-Lab/DriveHz}%
    \quad
    \raisebox{-0.06em}{\includegraphics[height=1em]{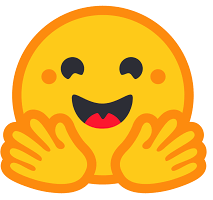}}\ Models:
    \href{https://huggingface.co/MxN0620/NAVSIM-Frequency-Checkpoints}{Policy \& Weights}%
  \end{adjustbox}%
  \par\vspace{0.35\baselineskip}%
  \manuscriptabscontent
}
\hypersetup{
  pdftitle={More Camera Frames, Better Driving? Within-Scene Information Scaling in Autonomous Driving},
  pdfauthor={Yumao Liu, Rui Qian, Qiang Huang, Jiahao Wei, Tao Liu, Xiangyu Li, Ke Ma}
}

\begin{document}
\begin{abstract}
Autonomous driving requires models to learn complex interactions among the ego vehicle, surrounding agents, and dynamic environments from continuously evolving driving processes. Existing work focuses on scaling model capacity and scaling training information exposure, largely treating driving scenes as the fundamental unit of training information. However, how the information exposed within each scene should scale with model capacity remains underexplored. We study \emph{within-scene information scaling}, using temporal sampling frequency as a controlled probe to vary camera-frame density within fixed driving scenes. We formulate a minimal information--capacity model that captures the interaction among bounded driving-relevant information, increasing driving-irrelevant exposure, and finite model capacity. The model characterizes the capacity-dependent scaling behavior of within-scene information and predicts that the preferred temporal sampling frequency increases with model capacity under its assumptions. We conduct controlled frequency sweeps across model families and driving datasets. These sweeps reveal complementary effects of model capacity and driving-irrelevant exposure. Within the E2EDriver family, increasing capacity shifts the preferred temporal sampling frequency for ADE/FDE upward, whereas increasing driving-irrelevant exposure through added noise at fixed capacity shifts it downward. These targeted interventions support the capacity-dependent prediction and finite-capacity competition in our minimal information--capacity model. Iteration-matched AutoVLA controls show that denser temporal sampling exposes additional driving-relevant information that additional optimization on sparse data cannot fully recover. Cross-model comparisons and structured-input HiVT further distinguish strong driving performance under sparse temporal exposure from gains through denser sampling. These findings motivate jointly developing model architectures, learning objectives, and within-scene data construction to utilize driving-relevant information while limiting driving-irrelevant exposure.
\end{abstract}
\maketitle

% ==============================================================
% 1. INTRODUCTION
% ==============================================================
\section{Introduction}
\label{sec:intro}

Autonomous driving requires models to learn complex interactions among the ego vehicle, surrounding agents, and dynamic environments from continuously evolving driving processes~\citep{chen2024end,Hu2023UniAD}. This requirement poses the core challenge for models to effectively extract and utilize information from large-scale driving scenes. Researchers identified two important research directions to address this challenge: scaling model capacity and scaling training information exposure~\citep{zhou2025autovla,li2025drivevla,dimlioglu2026scalingaware}. 

\begin{figure}[!t]
  \centering
  \includegraphics[width=\linewidth]{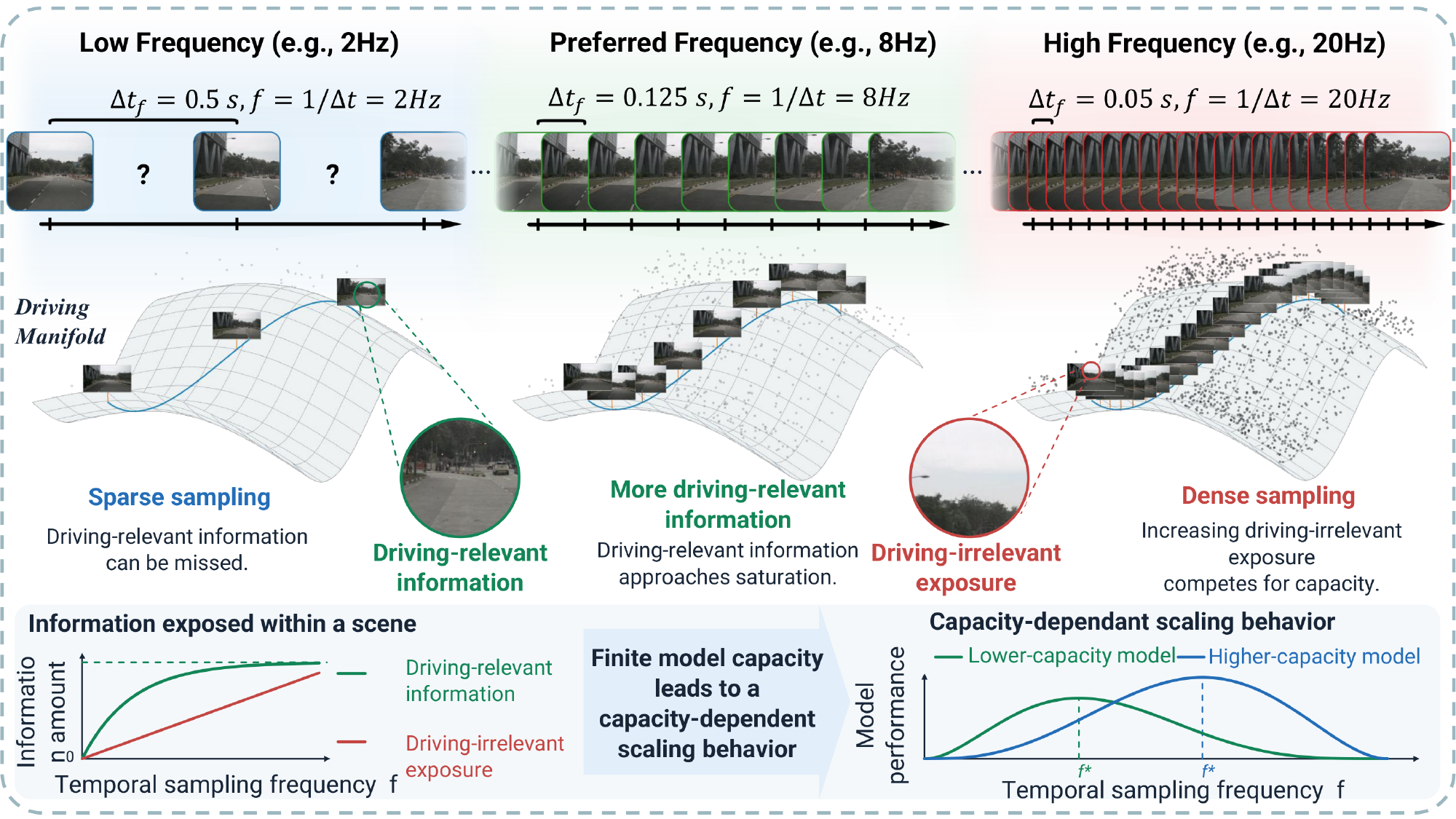}
  \vspace{-4pt}
  \caption{\textbf{Within-scene information scaling under finite model capacity.}
  Sparse temporal sampling can miss driving-relevant information. As temporal sampling frequency increases, driving-relevant information increases toward a finite limit, while driving-irrelevant exposure can continue increasing. Their interaction with finite model capacity shapes frequency-response curves, illustrating the capacity-dependent scaling behavior of within-scene information. Under the model assumptions, the preferred temporal sampling frequency increases with model capacity.}
  \label{fig:manifold_illustration}
  \vspace{-4pt}
\end{figure}

Existing works have explored these two scaling directions from complementary perspectives. On the model side, advances in spatiotemporal representation learning and unified driving architectures have improved how models represent and reason about driving scenes. BEVFormer develops spatiotemporal scene representations that aggregate observations across time~\citep{Li2022BEVFormer}, while UniAD integrates perception, prediction, and planning within a unified framework~\citep{Hu2023UniAD}. Recent vision-language-action and world-action models further extend this direction with larger-scale models and richer supervision~\citep{zhou2025autovla,li2025drivevla}. On the information side, existing approaches mainly focus on expanding driving data coverage and improving scene-level data composition. For example, MOSAIC selects informative driving scene clips based on their estimated marginal gains~\citep{dimlioglu2026scalingaware}. However, existing scaling strategies largely treat driving scenes as the fundamental unit of training information: they study which scenes should be included, how scenes should be represented, and how models should learn from these scenes, while leaving unexplored how the information contained within each individual scene should scale with model capacity.

In this work, we study a complementary scaling dimension: \emph{within-scene information scaling}. Beyond scaling the number of driving scenes, we ask: \emph{how should the information exposed within each scene scale with the model?} A driving scene is a continuous process. During data collection, this continuous process is discretized into camera frames with a certain temporal sampling frequency. Within-scene information scaling therefore corresponds to increasing the density of camera frames by increasing the temporal sampling frequency. These camera frames provide driving-relevant information, such as vehicle motion, agent interactions, and scene evolution, but also contain driving-irrelevant exposure~\citep{Feichtenhofer2019SlowFast,Hu2023UniAD}, including redundant content and visual variations unrelated to the driving decision~\citep{Ma2024SimCache}. 

\newpage
We hypothesize that increasing temporal sampling frequency changes the composition of information exposed to the model. At low temporal sampling frequency, sparse sampling can miss driving-relevant information such as temporal changes and interactions. As temporal sampling frequency increases, driving-relevant information increases but eventually approaches a finite limit, whereas driving-irrelevant exposure can continue increasing through redundant content and visual variations unrelated to the driving decision. Under finite model capacity, increasing driving-irrelevant exposure competes for capacity, so different models may utilize different amounts of within-scene information. We formulate a minimal information--capacity model that captures the interaction among driving-relevant information, driving-irrelevant exposure, and finite model capacity. The model characterizes a capacity-dependent scaling behavior of within-scene information. A concrete implication of this behavior is that the preferred temporal sampling frequency depends on model capacity. Figure~\ref{fig:manifold_illustration} illustrates this capacity-dependent scaling behavior by relating temporal sampling frequency to the changing composition of within-scene information and the resulting frequency-response curves across model capacities.

To empirically study this capacity-dependent scaling behavior, we use temporal sampling frequency as a controlled probe of within-scene information scaling. Within each frequency sweep, we fix the scene amount in the training set, model architecture, input construction, prediction target, and evaluation protocol, varying only temporal sampling frequency. We evaluate frequency-response curves across compact end-to-end models, large VLA models, and world-action models on Waymo~\citep{xu2025wod}, nuScenes~\citep{caesar2020nuscenes}, PAVE~\citep{li2025pave}, and in NAVSIM-based open-loop and closed-loop settings. We further conduct within-family capacity sweeps, targeted driving-irrelevant exposure interventions, and iteration-matched controls. The E2EDriver interventions show that the preferred temporal sampling frequency for ADE/FDE increases with model capacity but decreases with additional driving-irrelevant exposure introduced as noise, supporting the capacity-dependent prediction and finite-capacity competition in our minimal information--capacity model. Iteration-matched controls show that denser sampling exposes additional driving-relevant information for AutoVLA that repeated optimization on sparse data cannot fully recover, while other models demonstrate different abilities to utilize driving-relevant information under sparse and dense temporal exposure. These findings distinguish improvement from denser temporal exposure from driving performance already attainable under sparse exposure. They motivate evaluating both driving performance and frequency-response curves to guide the joint development of model architectures, learning objectives, and within-scene data construction, including how densely and which camera frames to sample.

\begingroup
% Keep the lead-in with the first item; allow breaks between contributions.
\makeatletter
\@beginparpenalty=10000
\makeatother
Our contributions are summarized as follows:
{\setlength{\leftmargini}{1.5em}\setlength{\topsep}{2pt}
\begin{itemize}
    \setlength{\itemsep}{0pt}
    \setlength{\parsep}{0pt}
    \item A within-scene information scaling perspective for autonomous driving models.
    We study how the information exposed within each driving scene should scale with model capacity, identifying a scaling dimension complementary to model capacity scaling and scene-level data scaling.

    \item A minimal information--capacity model for within-scene information scaling.
    We characterize the capacity-dependent scaling behavior of within-scene information through the interaction among driving-relevant information, driving-irrelevant exposure, and finite model capacity. Under the model assumptions, the preferred temporal sampling frequency increases with model capacity.

\item Controlled frequency-response evaluation and targeted tests of the information--capacity mechanism.
We use fixed-epoch sweeps to characterize model-dependent frequency responses and iteration-matched controls to show that denser sampling exposes driving-relevant information that repeated optimization on sparse data cannot fully recover. We further test the capacity-dependent prediction and finite-capacity competition through targeted capacity and driving-irrelevant exposure interventions.
\end{itemize}
}
\endgroup

%
% ==============================================================
% 2. RELATED WORK
% ==============================================================
\section{Related Work}
\label{sec:related}

\begingroup
\setlength{\parskip}{3pt}
\titlespacing{\paragraph}{0pt}{4pt}{1em}
\paragraph{Model capacity scaling and driving representations, architectures, and supervision.}
On the model-capacity-scaling side, existing works improve how driving models represent and reason about driving scenes through advances in spatiotemporal representations, unified driving architectures, and richer supervision. BEVFormer aggregates spatial and temporal observations into a bird's-eye-view representation~\citep{Li2022BEVFormer}. UniAD combines perception, prediction, and planning in a joint system~\citep{Hu2023UniAD}, while VAD and SparseDrive explore vectorized or sparse scene representations~\citep{Jiang2023VAD,Sun2024SparseDrive}. AutoVLA extends this direction through a vision-language-action formulation~\citep{zhou2025autovla}. DriveVLA-W0 adds future-image prediction to action supervision and studies how world modeling affects data scaling~\citep{li2025drivevla}.

\paragraph{Driving data coverage and scene-level data composition.}
Data scaling also depends on which driving scenes are included in training. MOSAIC estimates how clips from different driving domains contribute to evaluation metrics and allocates a collection budget accordingly~\citep{dimlioglu2026scalingaware}. Such scene-level selection is complementary to within-scene information scaling: scene-level scaling changes which driving scenes are included, whereas within-scene scaling changes the density of camera frames within fixed driving scenes.

\paragraph{Temporal information and redundancy.}
Camera frames provide driving-relevant information, such as vehicle motion, agent interactions, and scene evolution~\citep{Hu2023UniAD,Feichtenhofer2019SlowFast}. SlowFast uses pathways with different temporal rates for video recognition~\citep{Feichtenhofer2019SlowFast}. EVE avoids redundant computation across point-cloud frames~\citep{peng2021efficient}, and SimCache reuses VLM computations for similar scene regions~\citep{Ma2024SimCache}. These methods primarily study temporal representation or computational reuse rather than how the composition of information exposed within a fixed driving scene changes with temporal sampling frequency.

\paragraph{Within-scene information scaling and finite model capacity.}
Scaling-law studies relate performance to model size, data volume, and training compute~\citep{kaplan2020scaling,hoffmann2022training}. The Back to Basics study provides a different perspective by examining how learning data structure and high-dimensional noise place different demands on a model~\citep{Li2025B2B}. Motivated by these perspectives, our formulation models within-scene information scaling as an interaction among driving-relevant information, driving-irrelevant exposure, and finite model capacity, yielding a capacity-dependent preferred temporal sampling frequency.

\paragraph{Driving evaluation.}
Trajectory displacement errors measure agreement with recorded futures, but they do not capture every aspect of driving performance. NAVSIM closed-loop evaluations score predicted plans with simulation-based metrics that capture aspects of driving beyond displacement errors~\citep{dauner2024navsim,cao2025pseudosimulation}. We therefore report trajectory ADE/FDE together with NAVSIM EPDMS in the corresponding evaluation settings.
\par
\endgroup

%
% ==============================================================
% 3. INFORMATION--CAPACITY MODEL
% ==============================================================
\section{Within-Scene Information Scaling under Finite Model Capacity}
\label{sec:theory}

As temporal sampling frequency $f>0$ increases, driving-relevant information $\mathcal I(f)$ approaches a finite limit, whereas driving-irrelevant exposure $\mathcal B(f)$ can continue increasing. The interval between sampled camera frames within each driving scene is $\Delta t=1/f$. Within each frequency sweep, the scene amount in the training set, model architecture, input construction, prediction target, and evaluation protocol remain fixed. We model this distinction as

\begin{equation}
\mathcal I(f)=\Omega\frac{(f/f_0)^p}{1+(f/f_0)^p},
\qquad
\mathcal B(f)\approx\nu f,
\label{eq:core-information-exposure}
\end{equation}
where $\Omega,f_0,\nu>0$ and $p>1$. Here, $\Omega$ is the finite limit of driving-relevant information, half of which is exposed at $f_0$; $p$ controls the increase under sparse temporal sampling. Driving-irrelevant exposure includes redundant content and visual variations unrelated to the driving decision, with $\nu$ setting its linear growth rate. We use this linear approximation for the analysis. Both quantities are defined on a common normalized scale, not measured as Shannon information.

Let $\mathcal R(f):=\mathcal I(f)+\mathcal B(f)$ denote the combined modeled exposure and $C>0$ denote finite model capacity on the same normalized scale. We model the driving-relevant information effectively utilized through the normalized measure

\begin{equation}
\alpha_C(f)
:=\widetilde{\mathcal I}_C(f)
=\frac{\mathcal I(f)}{\mathcal R(f)}
 \frac{C}{C+\mathcal R(f)}
=\frac{C\,\mathcal I(f)}
 {\mathcal R(f)\bigl(C+\mathcal R(f)\bigr)}.
\label{eq:core-effective-information}
\end{equation}

The first factor measures the driving-relevant share of the combined modeled exposure. The second models the fraction that can be processed: it decreases with exposure and increases with model capacity. At fixed $C$, we assume that driving performance $\mathcal P_C(f)$, oriented so that higher is better, is a strictly increasing function of $\alpha_C(f)$. The preferred temporal sampling frequency therefore satisfies

\begin{equation}
f_C^\star\in\underset{f>0}{\arg\max}\;\mathcal P_C(f)
=\underset{f>0}{\arg\max}\;\alpha_C(f).
\label{eq:core-optimum}
\end{equation}

For $p>1$, $\alpha_C(f)$ approaches zero at both frequency extremes: sparse temporal sampling misses driving-relevant information, while very dense temporal sampling adds driving-irrelevant exposure after driving-relevant information has nearly saturated. The model therefore has a finite interior preferred temporal sampling frequency. Holding the fixed driving scenes and the parameters $\Omega,f_0,p,\nu$ unchanged, a nondegenerate interior optimum that varies smoothly with $C$ satisfies

\begin{equation}
\frac{\mathrm d f_C^\star}{\mathrm d C}>0.
\label{eq:core-capacity-monotonicity}
\end{equation}

Thus, greater model capacity allows denser temporal exposure to be processed, shifting the preferred temporal sampling frequency upward under the model assumptions. On a finite tested frequency set, this frequency can remain unchanged or increase. This shift is one consequence of the capacity-dependent scaling behavior of within-scene information. \Cref{app:capacity_derivation} provides the derivation.

% ==============================================================
% 4. EXPERIMENTS
% ==============================================================
\section{Experiments}
\label{sec:experiments}

\begin{figure}[tb]
  \centering
  \includegraphics[width=\linewidth]{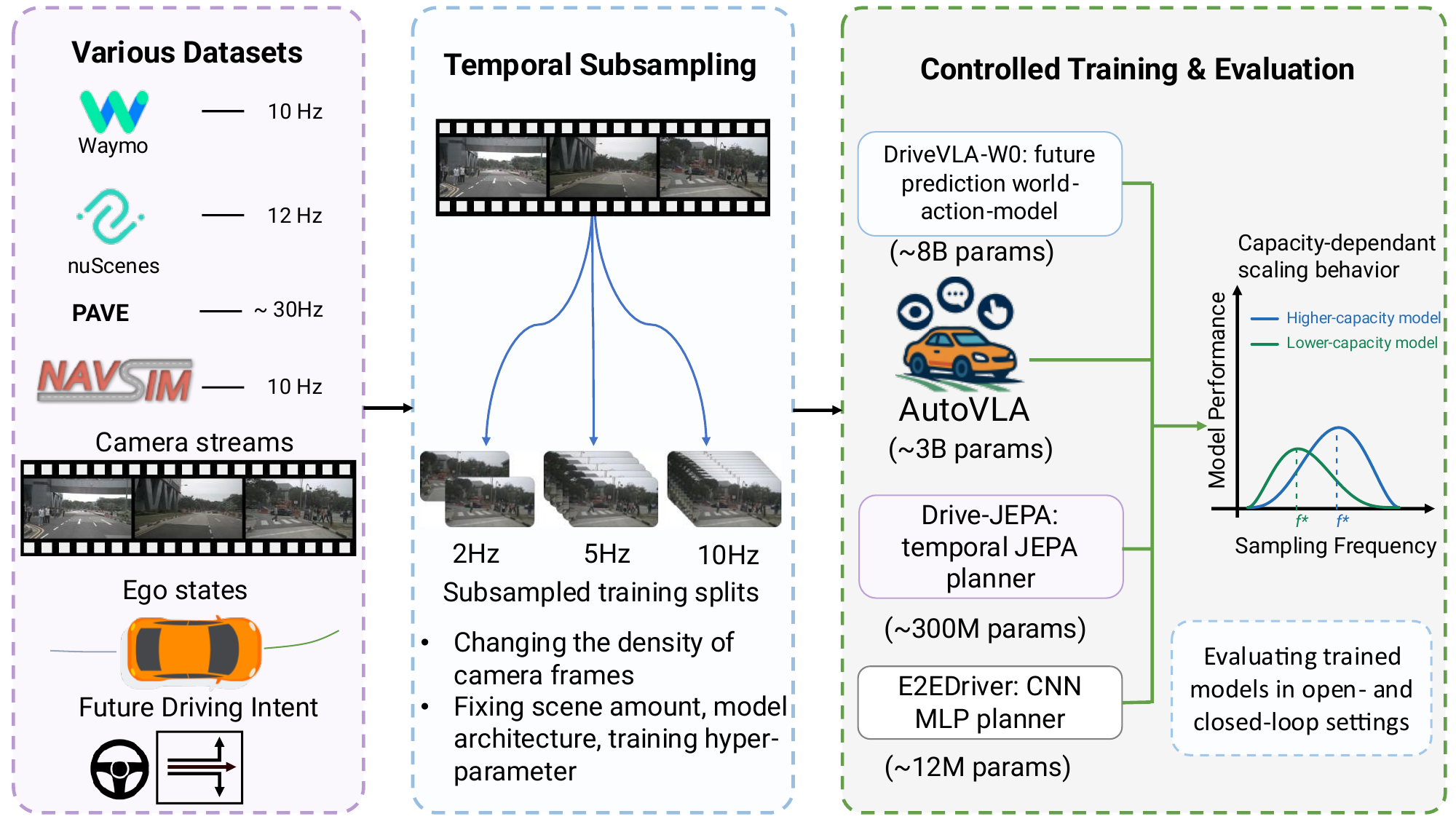}
    \caption{\textbf{Temporal sampling frequency as a controlled probe of within-scene information scaling.}
    Training sets contain different densities of sampled camera frames from the same driving scenes, while model architecture, input construction, prediction target, and evaluation protocol remain fixed within each frequency sweep.
    The resulting frequency-response curves characterize how different models utilize increasingly dense temporal exposure.}
  \label{fig:experiment_pipeline}
\end{figure}

We use temporal sampling frequency as a controlled probe of within-scene information scaling across compact end-to-end models, large VLA models, and world-action models. We analyze frequency-response curves to examine both driving performance under sparse temporal exposure and its changes as temporal sampling becomes denser. Iteration-matched comparisons assess the effect of the number of optimization updates on these responses. \Cref{fig:experiment_pipeline} illustrates the training-set construction for the open-loop and NAVSIM experiments. Dataset-handling details are provided in \Cref{app:timestamp_anchor}.

\paragraph{Task and metrics.}
We study command-conditioned ego-trajectory prediction~\citep{codevilla2018end}. At timestamp $t$, each model receives model-specific observations and predicts a future ego trajectory. The open-loop sweeps report 3-second ADE and FDE in meters~\citep{ettinger2021large}, with 5-second results where available. The NAVSIM closed-loop evaluations report EPDMS as a percentage~\citep{dauner2024navsim,cao2025pseudosimulation} alongside 3-second ADE/FDE. On NAVTEST, ADE averages positional error at six targets from 0.5 to 3.0 seconds, and FDE uses the 3.0-second target. EPDMS includes all 12,146 scenarios; ADE/FDE use valid trajectories only. Higher EPDMS and lower ADE/FDE indicate better driving performance under their respective evaluation protocols. Throughout the empirical comparisons, preferred temporal sampling frequency refers to a tested frequency attaining the best reported mean for the stated metric and evaluation protocol. ADE/FDE characterize trajectory prediction, whereas EPDMS characterizes NAVTEST driving performance, with each metric determining its own preferred frequency. This interpretation applies consistently across model families.

\paragraph{Datasets.}
The open-loop sweeps use Waymo~\citep{xu2025wod}, nuScenes~\citep{caesar2020nuscenes}, and PAVE~\citep{li2025pave}, with highest evaluated temporal sampling frequencies of 10\,Hz, 12\,Hz, and 20\,Hz, respectively. The PAVE sweep stops below its approximately 30\,Hz native camera frequency. \Cref{tab:datasets} summarizes the frequency ranges, driving domains, and training-set statistics. NAVSIM-10Hz uses native camera frames from the corresponding nuPlan recordings to construct a 10\,Hz version of the official 2\,Hz NAVSIM training clips within the same driving scenes~\citep{caesar2021nuplan,dauner2024navsim}.

\begin{table}[!htbp]
\caption{Dataset summary and training-set motion statistics. Mean ego speed and displacement per sampled camera frame are computed at each dataset's highest evaluated temporal sampling frequency. NAVSIM-10Hz uses native nuPlan camera frames within the NAVTRAIN driving scenes, not upsampled OpenScene images.}
  \label{tab:datasets}
  \centering
  \renewcommand{\arraystretch}{1.08}
  \setlength{\tabcolsep}{5pt}
  \begin{adjustbox}{max width=\linewidth}
    \begin{tabular}{lcccc}
    \specialrule{1.2pt}{0pt}{0pt}
    \rowcolor[gray]{.90} 
     & \textbf{Waymo} & \textbf{nuScenes} & \textbf{PAVE} & \textbf{NAVSIM-10Hz} \\
    \midrule
    Camera-frame frequency & 10\,Hz & 12\,Hz & ${\sim}$30\,Hz & 10\,Hz \\
    Highest evaluated f & 10\,Hz & 12\,Hz & 20\,Hz & 10\,Hz \\
    Cameras & 8 & 6 & 4 & 8 \\
    Dominant scenario & Urban/suburban & Urban & Highway/expressway & Urban/suburban \\
    Training samples & ${\sim}$365k & ${\sim}$116k & ${\sim}$167k & ${\sim}$425k \\
    Mean ego speed (m/s) & 5.80 & 5.05 & 13.75 & 4.67 \\
    Displacement/frame (m) & 0.58 & 0.42 & 0.69 & 0.46 \\
    Source & Public & Public & Public & Public \\
    \specialrule{1.2pt}{0pt}{0pt}
  \end{tabular}
    \end{adjustbox}
\end{table}

\paragraph{Models and training.}
The open-loop sweeps evaluate E2EDriver, BEV-E2EDriver, Tiny-SSR, and AutoVLA. The three compact end-to-end models have approximately 10--18M parameters and use CNN-based multi-view image encoding, a bird's-eye-view representation, and a compact sparse-scene representation, respectively. AutoVLA uses a roughly 3B-parameter vision-language-action backbone to predict trajectory action tokens~\citep{zhou2025autovla}. DriveVLA-W0~\citep{li2025drivevla}, which uses world-model supervision, is additionally evaluated on nuScenes. The NAVTEST evaluations compare Drive-JEPA (pose-only)~\citep{wang2026drivejepa}, a temporal JEPA planner, with E2EDriver, AutoVLA (answer-only SFT), and DriveVLA-W0 (Emu3Pi0). For the NAVTEST results, we use Drive-JEPA, AutoVLA, and DriveVLA-W0 as shorthand for these evaluated configurations; the NAVTEST tables retain their full configuration names. Within each frequency sweep, model architecture, input construction (including the command representation), and prediction target remain fixed. Temporal sampling frequency therefore controls the density of sampled camera frames within fixed driving scenes, rather than the temporal context within each input. Detailed model and loss descriptions for the open-loop and NAVTEST sweeps are provided in \Cref{app:model_loss_details}. The open-loop sweeps keep the epoch count fixed within each model--dataset pair. Increasing temporal sampling frequency therefore increases the number of sampled camera frames per epoch and the number of optimization updates, motivating the iteration-matched comparisons below. For the NAVTEST sweeps, Drive-JEPA, E2EDriver, AutoVLA, and DriveVLA-W0 use 20, 10, 5, and 4 epochs, respectively, at every temporal sampling frequency. We report the final-epoch checkpoint without selecting settings based on performance. Higher temporal sampling frequencies therefore receive more optimization updates in the fixed-epoch sweeps. The epoch counts and checkpoint policy are summarized in \Cref{tab:navtest_fixed_epochs} in \Cref{app:model_loss_details}.

\begin{figure}[tb]
  \centering
  \includegraphics[width=\linewidth]{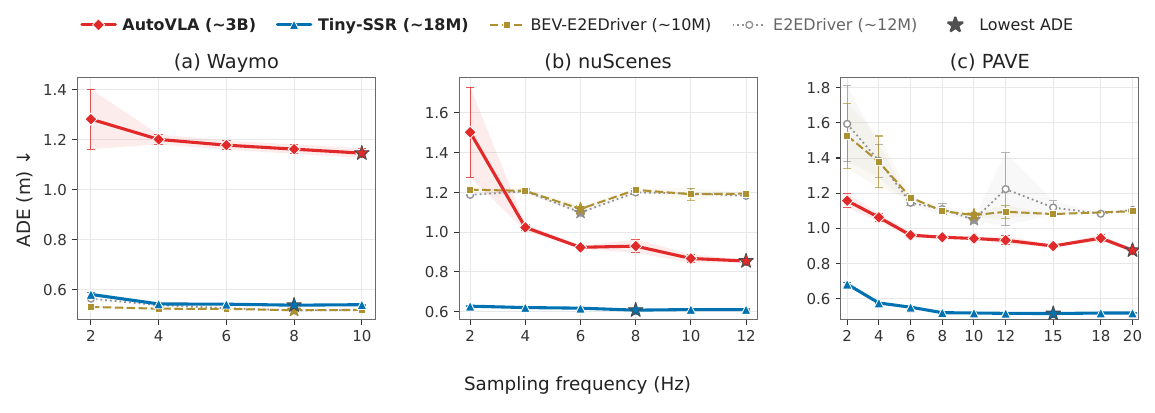}
  \caption{\textbf{Open-loop frequency-response curves across models and datasets.}
  Compact end-to-end models show intermediate ADE minima or near-plateau regions, whereas AutoVLA reaches its lowest reported ADE at the highest tested temporal sampling frequency on each dataset.}
  \label{fig:freq_response}
\end{figure}

\paragraph{Open-loop frequency-response curves depend on model and dataset.}
\Cref{tab:future_3s} reports the 3-second open-loop sweeps, and \Cref{fig:freq_response} summarizes the corresponding ADE frequency-response curves. Compact end-to-end models often show intermediate minima or near-plateau regions, whereas AutoVLA reaches its lowest reported ADE at the highest tested temporal sampling frequency on each dataset. This contrast shows that increasing the density of sampled camera frames does not produce uniform improvements across models. Importantly, the improvement from denser temporal exposure and the level of driving performance are distinct: Tiny-SSR at 2\,Hz already has lower ADE than AutoVLA at its highest tested frequency on all three datasets. A higher preferred temporal sampling frequency within the tested range is therefore not, by itself, evidence of better driving performance. The repeated runs also distinguish different frequency-response shapes: E2EDriver has a pronounced intermediate ADE minimum on nuScenes but a near-plateau region on Waymo. Near such a plateau, the range of temporal sampling frequencies with similar performance is more informative than the frequency with the lowest reported mean. These observations motivate studying within-scene information scaling jointly with model architecture and the driving scenes, using frequency-response curves to assess performance under sparse temporal exposure, changes under denser temporal exposure, and the reported run-to-run variation.

\begin{table}[!t]
\caption{\textbf{Three-second open-loop trajectory prediction across temporal sampling frequencies.} ADE/FDE are in meters. ADE is reported as mean~$\pm$~standard deviation over three runs where available. Bold marks the lowest mean ADE for each model--dataset pair and the FDE at that setting. Some high-frequency results are grouped into ranges.}
    \label{tab:future_3s}
    \centering
  \renewcommand{\arraystretch}{1.08}
    \small
    \begin{adjustbox}{max width=\linewidth}
    \begin{tabular}{ll|cc|cc|cc}
        \specialrule{1.2pt}{0pt}{0pt}
        \rowcolor[gray]{.90} 
     &  & \multicolumn{2}{c|}{\textbf{Waymo} (10\,Hz)} & \multicolumn{2}{c|}{\textbf{nuScenes} (12\,Hz)} & \multicolumn{2}{c}{\textbf{PAVE} (20\,Hz eval.)} \\
        \cmidrule(lr){3-4} \cmidrule(lr){5-6} \cmidrule(lr){7-8}
        \textbf{Model} & \textbf{Freq.} & \textbf{ADE$\downarrow$} & \textbf{FDE$\downarrow$} & \textbf{ADE$\downarrow$} & \textbf{FDE$\downarrow$} & \textbf{ADE$\downarrow$} & \textbf{FDE$\downarrow$} \\
        \midrule
        \multirow{8}{*}{E2EDriver}
        & 2\,Hz  & 0.564\std{.002} & 1.584 & 1.188\std{.004} & 3.082 & 1.594\std{.216} & 4.123 \\
        & 4\,Hz  & 0.538\std{.001} & 1.527 & 1.206\std{.011} & 3.150 & 1.384\std{.093} & 3.716 \\
        & 6\,Hz  & 0.526\std{.002} & 1.500 & \textbf{1.099}\std{.006} & \textbf{2.856} & 1.145\std{.016} & 2.523 \\
        & 8\,Hz  & \textbf{0.517}\std{.002} & \textbf{1.468} & 1.200\std{.005} & 3.134 & 1.109\std{.031} & 2.478 \\
        & 10\,Hz & 0.518\std{.003} & 1.469 & 1.195\std{.002} & 3.114 & \textbf{1.049}\std{.007} & \textbf{2.397} \\
        & 12\,Hz & --- & --- & 1.183\std{.007} & 3.093 & 1.223\std{.208} & 2.666 \\
        & 15\,Hz & --- & --- & --- & --- & 1.118\std{.039} & 2.437 \\
        & 18--20\,Hz & --- & --- & --- & --- & 1.083--1.105 & 2.374--2.478 \\
        \midrule
        \multirow{7}{*}{\shortstack[l]{BEV-\\E2EDriver}}
        & 2\,Hz  & 0.531\std{.003} & 1.485 & 1.214\std{.012} & 3.137 & 1.527\std{.185} & 3.952 \\
        & 4\,Hz  & 0.524\std{.003} & 1.479 & 1.208\std{.012} & 3.121 & 1.378\std{.147} & 3.629 \\
        & 6\,Hz  & 0.523\std{.003} & 1.491 & \textbf{1.117}\std{.016} & \textbf{2.875} & 1.175\std{.010} & 2.588 \\
        & 8\,Hz  & \textbf{0.518}\std{.001} & \textbf{1.464} & 1.213\std{.006} & 3.126 & 1.098\std{.003} & 2.491 \\
        & 10\,Hz & 0.519\std{.003} & 1.464 & 1.191\std{.031} & 3.105 & \textbf{1.075}\std{.014} & \textbf{2.422} \\
        & 12\,Hz & --- & --- & 1.194\std{.004} & 3.094 & 1.094\std{.036} & 2.478 \\
        & 15--20\,Hz & --- & --- & --- & --- & 1.081--1.097 & 2.423--2.435 \\
        \midrule
        \multirow{8}{*}{Tiny-SSR}
        & 2\,Hz  & 0.581\std{.008} & 1.443 & 0.628\std{.001} & 1.465 & 0.682\std{.011} & 1.587 \\
        & 4\,Hz  & 0.543\std{.003} & 1.348 & 0.621\std{.002} & 1.459 & 0.575\std{.005} & 1.412 \\
        & 6\,Hz  & 0.542\std{.001} & 1.347 & 0.618\std{.002} & 1.455 & 0.551\std{.004} & 1.159 \\
        & 8\,Hz  & \textbf{0.538}\std{.002} & \textbf{1.334} & \textbf{0.608}\std{.001} & \textbf{1.440} & 0.520\std{.005} & 1.049 \\
        & 10\,Hz & 0.540\std{.002} & 1.340 & 0.611\std{.003} & 1.447 & 0.518\std{.003} & 1.049 \\
        & 12\,Hz & --- & --- & 0.611\std{.003} & 1.448 & 0.516\std{.005} & 1.052 \\
        & 15\,Hz & --- & --- & --- & --- & \textbf{0.515}\std{.003} & \textbf{1.047} \\
        & 18--20\,Hz & --- & --- & --- & --- & 0.518\std{.002} & 1.062 \\
        \midrule
        \multirow{9}{*}{AutoVLA}
        & 2\,Hz  & 1.282\std{.120} & 2.881 & 1.503\std{.227} & 3.331 & 1.157\std{.040} & 2.450 \\
        & 4\,Hz  & 1.201\std{.020} & 2.755 & 1.024\std{.016} & 2.398 & 1.062\std{.022} & 2.200 \\
        & 6\,Hz  & 1.178\std{.018} & 2.684 & 0.924\std{.007} & 2.184 & 0.961\std{.015} & 2.016 \\
        & 8\,Hz  & 1.162\std{.019} & 2.718 & 0.930\std{.034} & 2.176 & 0.949\std{.005} & 1.973 \\
        & 10\,Hz & \textbf{1.146}\std{.019} & \textbf{2.635} & 0.868\std{.021} & 2.040 & 0.942\std{.002} & 1.930 \\
        & 12\,Hz & --- & --- & \textbf{0.855}\std{.008} & \textbf{2.029} & 0.932\std{.025} & 1.907 \\
        & 15\,Hz & --- & --- & --- & --- & 0.899\std{.004} & 1.844 \\
        & 18\,Hz & --- & --- & --- & --- & 0.944\std{.019} & 1.941 \\
        & 20\,Hz & --- & --- & --- & --- & \textbf{0.875}\std{.004} & \textbf{1.823} \\
        \specialrule{1.2pt}{0pt}{0pt}
    \end{tabular}
    \end{adjustbox}
\end{table}

\begingroup
\setlength{\parskip}{3pt}
\titlespacing{\paragraph}{0pt}{4pt}{1em}
\paragraph{Within-family E2EDriver capacity sweep.}
We vary E2EDriver's internal width while keeping the pretrained visual backbone, input construction, training objectives, and training settings at each temporal sampling frequency fixed. The lowest mean ADE/FDE shifts from 6\,Hz to 8\,Hz to 10\,Hz as internal width increases (\Cref{tab:e2edriver_width_configs,tab:e2edriver_width_sweep} in \Cref{app:model_loss_details}). Our controlled E2EDriver experiments support a central prediction of our minimal information--capacity model: the preferred temporal sampling frequency for ADE/FDE can increase with model capacity when other model and training settings are held fixed. For Tiny E2EDriver, 10\,Hz improves EPDMS by 1.97 percentage points over 6\,Hz. Denser temporal sampling can therefore improve driving performance even when trajectory errors have nearly saturated.

\paragraph{Iteration-matched controls.}
We assess whether the frequency-response curves can be explained by the number of optimization updates alone. We pair lower temporal sampling frequencies with more epochs and higher frequencies with fewer epochs, using frequency $\times$ epochs to approximately match the number of optimization updates. The comparisons in \Cref{app:iteration_matched} and \Cref{tab:iteration_matched} show that lower-frequency configurations retain comparable or lower ADE after this adjustment. For E2EDriver, 6\,Hz for 10 epochs has 12.05\% lower ADE than 12\,Hz for 5 epochs on nuScenes, whereas the analogous difference on Waymo is only 0.12\%. Increasing epochs revisits sampled camera frames, whereas increasing temporal sampling frequency changes within-scene information exposure. The results support treating within-scene information scaling and the number of optimization updates as distinct training choices: denser temporal exposure does not necessarily improve driving performance even when the update counts are approximately matched.

\paragraph{5-second results.}
The 5-second evaluations in \Cref{app:five_second} (\Cref{tab:future_5s}) retain the model- and dataset-dependent frequency-response patterns, while the preferred temporal sampling frequency can change with the prediction target. This motivates evaluating within-scene information scaling for the prediction target rather than assigning each model a fixed preferred temporal sampling frequency.

\paragraph{DriveVLA-W0 frequency response on nuScenes.}
\Cref{tab:drivevla_nuscenes} shows that most of DriveVLA-W0's ADE reduction occurs between 2\,Hz and 4\,Hz. Increasing temporal sampling frequency from 4\,Hz to 12\,Hz further reduces its mean ADE by approximately 1.6\%, compared with 16.5\% for AutoVLA on nuScenes (\Cref{tab:future_3s}). Although both models reach their lowest reported ADE at 12\,Hz, the extent of improvement from denser temporal exposure differs substantially. The frequency-response curve is therefore more informative for within-scene data construction than the best tested frequency alone. These results motivate studying how temporal sampling can expose additional driving-relevant information that affects the driving decision, rather than assuming that VLA and world-action models all require uniformly dense temporal exposure.
\par
\endgroup

\begin{table}[!htbp]
\caption{\textbf{DriveVLA-W0 (Emu3Pi0) frequency response on nuScenes.} ADE/FDE are in meters; lower is better. Bold marks the lowest reported mean for each metric.}
  \label{tab:drivevla_nuscenes}
  \centering
  \renewcommand{\arraystretch}{1.08}
  \begin{adjustbox}{max width=\linewidth}
    \begin{tabular}{ccc}
    \specialrule{1.2pt}{0pt}{0pt}
    \rowcolor[gray]{.90} 
    \textbf{Training frequency} & \textbf{ADE (m) $\downarrow$} & \textbf{FDE (m) $\downarrow$} \\
    \midrule
    2\,Hz  & 1.321$\pm$0.002 & 2.964$\pm$0.003 \\
    4\,Hz  & 1.256$\pm$0.001 & 2.915$\pm$0.005 \\
    6\,Hz  & 1.265$\pm$0.000 & 2.919$\pm$0.002 \\
    8\,Hz  & 1.258$\pm$0.002 & 2.922$\pm$0.004 \\
    10\,Hz & 1.242$\pm$0.003 & 2.872$\pm$0.006 \\
    12\,Hz & \textbf{1.236}$\pm$0.004 & \textbf{2.856}$\pm$0.006 \\
    \specialrule{1.2pt}{0pt}{0pt}
  \end{tabular}
    \end{adjustbox}
\end{table}

\paragraph{NAVTEST frequency-response curves and adaptive temporal sampling.}
\Cref{tab:navtest_frequency_response} reports NAVSIM closed-loop evaluations on NAVTEST over three random seeds per setting. Drive-JEPA and E2EDriver use uniform temporal sampling at 2, 4, 6, 8, and 10\,Hz; AutoVLA and DriveVLA-W0 are evaluated with uniform 2, 4, 6, 8, and 10\,Hz training and motion-based within-scene adaptive 4\,Hz training. The adaptive setting uses motion analysis to vary the sampling frequency over time within each scene, sampling more densely during highly dynamic periods and more sparsely when motion changes slowly. We distinguish the uniform frequency comparisons from adaptive temporal sampling, which also changes how camera frames are sampled. The A4 sampling procedure is provided in \Cref{app:timestamp_anchor}.

\begin{table}[!htbp]
\caption{\textbf{NAVTEST evaluation with uniform and adaptive temporal sampling.} EPDMS is in percent and includes all 12,146 NAVTEST scenarios; 3-second ADE/FDE are in meters and use valid trajectories. A4 denotes adaptive 4\,Hz training; all other settings use uniform temporal sampling. Bold marks the best reported mean within each model and metric.}
  \label{tab:navtest_frequency_response}
  \centering
  \renewcommand{\arraystretch}{1.08}
  \footnotesize
  \setlength{\tabcolsep}{3pt}
  \begin{adjustbox}{max width=\linewidth}
    \begin{tabular}{llccc}
    \specialrule{1.2pt}{0pt}{0pt}
    \rowcolor[gray]{.90} 
    \textbf{Model} & \textbf{Training frequency} & \textbf{EPDMS (\%) $\uparrow$} & \textbf{3-s ADE (m) $\downarrow$} & \textbf{3-s FDE (m) $\downarrow$} \\
    \midrule
    \multirow{5}{*}{Drive-JEPA (pose-only)}
      & 2\,Hz  & 89.554$\pm$0.085 & \textbf{0.350}$\pm$0.002 & \textbf{0.811}$\pm$0.010 \\
      & 4\,Hz  & \textbf{89.958}$\pm$0.068 & 0.389$\pm$0.003 & 0.878$\pm$0.005 \\
      & 6\,Hz  & 89.548$\pm$0.062 & 0.370$\pm$0.005 & 0.850$\pm$0.004 \\
      & 8\,Hz  & 89.608$\pm$0.016 & 0.378$\pm$0.001 & 0.881$\pm$0.004 \\
      & 10\,Hz & 89.597$\pm$0.119 & 0.386$\pm$0.002 & 0.883$\pm$0.006 \\
    \midrule
    \multirow{6}{*}{DriveVLA-W0 (Emu3Pi0)}
      & 2\,Hz  & 85.673$\pm$0.078 & 0.621$\pm$0.001 & 1.236$\pm$0.003 \\
      & 4\,Hz  & 86.5926$\pm$0.061 & 0.5848$\pm$0.001 & 1.1549$\pm$0.002 \\
      & 6\,Hz  & 87.0347$\pm$0.054 & 0.6058$\pm$0.002 & 1.1933$\pm$0.003 \\
      & 8\,Hz  & 87.3296$\pm$0.049 & 0.6364$\pm$0.001 & 1.2369$\pm$0.002 \\
      & A4 & 86.663$\pm$0.127 & \textbf{0.467}$\pm$0.001 & \textbf{1.016}$\pm$0.002 \\
      & 10\,Hz & \textbf{87.366}$\pm$0.093 & 0.6194$\pm$0.003 & 1.2053$\pm$0.005 \\
    \midrule
    \multirow{6}{*}{AutoVLA (answer-only SFT)}
      & 2\,Hz  & 78.793$\pm$0.062 & 0.506$\pm$0.000 & 1.150$\pm$0.001 \\
      & 4\,Hz  & 80.7333$\pm$0.0513 & 0.4827$\pm$0.0002 & \textbf{1.0917}$\pm$0.0003 \\
      & 6\,Hz  & 80.6389$\pm$0.0262 & \textbf{0.4818}$\pm$0.0001 & 1.0966$\pm$0.0006 \\
      & 8\,Hz  & 80.9534$\pm$0.0208 & 0.4831$\pm$0.0002 & 1.0987$\pm$0.0005 \\
      & A4 & 80.319$\pm$0.022 & 0.484$\pm$0.000 & 1.102$\pm$0.001 \\
      & 10\,Hz & \textbf{81.135}$\pm$0.085 & 0.485$\pm$0.000 & 1.103$\pm$0.001 \\
    \midrule
    \multirow{5}{*}{E2EDriver}
      & 2\,Hz  & 76.888$\pm$0.298 & 0.395$\pm$0.014 & 0.919$\pm$0.013 \\
      & 4\,Hz  & 79.138$\pm$0.308 & 0.382$\pm$0.006 & 0.893$\pm$0.005 \\
      & 6\,Hz  & 80.079$\pm$0.570 & 0.391$\pm$0.024 & 0.898$\pm$0.030 \\
      & 8\,Hz  & \textbf{81.558}$\pm$0.078 & \textbf{0.371}$\pm$0.001 & \textbf{0.876}$\pm$0.006 \\
      & 10\,Hz & 81.343$\pm$0.059 & 0.373$\pm$0.008 & 0.879$\pm$0.007 \\
    \specialrule{1.2pt}{0pt}{0pt}
  \end{tabular}
    \end{adjustbox}
\end{table}

\newpage
AutoVLA and DriveVLA-W0 achieve their highest mean EPDMS at uniform 10\,Hz among the evaluated sampling settings, including adaptive 4\,Hz (\Cref{tab:navtest_frequency_response}). In contrast, Drive-JEPA at 2\,Hz already achieves higher mean EPDMS and lower mean ADE/FDE than both models at uniform 10\,Hz. This distinguishes achieving strong driving performance under sparse temporal exposure from gaining performance through denser temporal exposure. The need for denser temporal sampling should therefore be evaluated through frequency-response curves rather than inferred from model capacity alone. Adaptive temporal sampling offers a complementary direction for data construction: studying which camera frames to expose, not only how densely to sample.

\paragraph{Large-model iteration-matched controls.}
We match optimization updates for AutoVLA and DriveVLA-W0 on NAVTEST (\Cref{tab:large_model_iteration_matched}). With optimization updates matched, 10\,Hz still improves AutoVLA EPDMS by 3.25 percentage points, with lower mean ADE/FDE. AutoVLA shows that denser temporal sampling exposes additional driving-relevant information. Additional optimization updates on sparse 2\,Hz data cannot fully recover this benefit. In contrast, DriveVLA-W0 recovers much of its high-frequency EPDMS gap through additional training on sparse anchors. For some models, such as DriveVLA-W0, repeated optimization on sparse, task-relevant data can improve driving performance. This repeated optimization is distinct from additional driving-irrelevant exposure. Its architecture and world-model pretraining may contribute to this behavior.

\begin{table}[!htb]
  \caption{\textbf{Large-model iteration-matched controls on NAVTEST.} Results are mean $\pm$ sample standard deviation over three random seeds. Bold marks the better mean for each metric within each model.}
  \label{tab:large_model_iteration_matched}
  \centering
  \renewcommand{\arraystretch}{1.08}
  \small
  \begin{adjustbox}{max width=\linewidth}
    \begin{tabular}{lcccc}
    \specialrule{1.2pt}{0pt}{0pt}
    \rowcolor[gray]{.90} 
    \textbf{Training frequency} & \textbf{Epochs} & \textbf{EPDMS (\%) $\uparrow$} & \textbf{3-s ADE (m) $\downarrow$} & \textbf{3-s FDE (m) $\downarrow$} \\
    \midrule
    \multicolumn{5}{@{}l}{\textbf{AutoVLA (answer-only SFT)}; batch size: 32} \\
    2\,Hz & 25 & 77.881$\pm$0.055 & 0.515959$\pm$0.000575 & 1.166427$\pm$0.001166 \\
    \rowcolor[gray]{.92} 10\,Hz & 5 & \textbf{81.135}$\pm$0.085 & \textbf{0.485}$\pm$0.000 & \textbf{1.103}$\pm$0.001 \\
    \midrule
    \multicolumn{5}{@{}l}{\textbf{DriveVLA-W0 (Emu3Pi0)}; batch size: 96} \\
    2\,Hz & 20 & 86.787$\pm$0.051 & 0.6198$\pm$0.0012 & \textbf{1.2016}$\pm$0.0009 \\
    \rowcolor[gray]{.92} 10\,Hz & 4 & \textbf{87.366}$\pm$0.093 & \textbf{0.6194}$\pm$0.003 & 1.2053$\pm$0.005 \\
    \specialrule{1.2pt}{0pt}{0pt}
  \end{tabular}
    \end{adjustbox}
\end{table}

\paragraph{Structured-input experiment with HiVT.}
Complementarily, the compact HiVT model~\citep{Zhou2022HiVT} achieves its lowest tested ADE/FDE at 10\,Hz when driving-relevant information is directly provided through structured inputs (\Cref{tab:hivt_structured_sweep} in \Cref{app:model_loss_details}). This comparison uses fixed-epoch frequency sweeps with eight epochs at each temporal sampling frequency. This motivates developing input representations that help models utilize driving-relevant information more effectively.

\paragraph{Targeted driving-irrelevant exposure intervention.}
We introduce additional driving-irrelevant exposure into the training samples of Original E2EDriver by adding noise first to camera images only, and then to both camera images and ego-state history. Compared with the original E2EDriver training, the frequencies yielding the lowest mean ADE/FDE shift from 8\,Hz to 6\,Hz and 4\,Hz, respectively (\Cref{tab:e2edriver_driving_irrelevant_noise} in \Cref{app:model_loss_details}). Additional driving-irrelevant exposure introduced as noise shifts the preferred temporal sampling frequency for ADE/FDE downward under fixed model capacity. This targeted intervention supports the finite-capacity competition in our minimal information--capacity model: driving-irrelevant exposure reduces the benefit of denser temporal sampling.

\newpage
\paragraph{Discussion.}
The controlled E2EDriver width sweep supports the prediction that the preferred temporal sampling frequency for ADE/FDE increases with model capacity. Additional driving-irrelevant exposure introduced as noise shifts the preferred temporal sampling frequency for ADE/FDE downward under fixed model capacity. This targeted intervention supports the finite-capacity competition in our minimal information--capacity model: driving-irrelevant exposure reduces the benefit of denser temporal sampling. AutoVLA shows that denser temporal sampling exposes additional driving-relevant information. Additional optimization updates on sparse 2\,Hz data cannot fully recover this benefit. DriveVLA-W0 recovers much of its high-frequency gap through longer sparse-anchor training, while Drive-JEPA performs strongly under sparse exposure and compact structured-input HiVT benefits from denser sampling. This contrast separates a model's performance under sparse exposure from its gain through densification. Frequency-response curves and driving performance should thus guide model architectures, learning objectives, and sampling policies. Adaptive temporal sampling offers a complementary direction for data construction: studying which camera frames to expose, not only how densely to sample, with choices evaluated for each model and driving-performance metric. The objective is to achieve strong driving performance when sparse temporal exposure contains sufficient driving-relevant information, exploit additional driving-relevant information when denser temporal sampling reveals it, and limit the influence of driving-irrelevant exposure.

%
% ==============================================================
% 6. CONCLUSION
% ==============================================================
\section{Conclusion}
\label{sec:conclusion}
We study within-scene information scaling in autonomous driving using temporal sampling frequency as a controlled probe. Our minimal information--capacity model characterizes the capacity-dependent scaling behavior of within-scene information through the interaction among driving-relevant information, driving-irrelevant exposure, and finite model capacity. Controlled E2EDriver width and driving-irrelevant exposure interventions provide complementary support for the model's capacity-dependent prediction and finite-capacity competition. Iteration-matched AutoVLA controls establish that denser temporal sampling exposes additional driving-relevant information that repeated optimization on sparse anchors cannot fully recover. Comparisons across model families and structured input representations show that strong driving performance under sparse exposure is distinct from improvements gained by denser exposure. Together, these results show why the preferred temporal sampling frequency alone cannot characterize a model's driving capability or its need for additional frames. Driving performance and frequency-response curves should jointly guide model architectures, learning objectives, and within-scene data construction. Within-scene information scaling is therefore not about maximizing camera-frame density, but about matching driving-relevant information exposure to model capacity and effective information utilization while limiting driving-irrelevant exposure.

\subsection*{AI use statement}
In preparing this manuscript, generative AI tools were used to assist with grammar checking and text polishing, LaTeX
formatting. All AI-assisted changes were reviewed by the authors for technical accuracy and consistency with the reported work. The authors take responsibility for the final content of the manuscript, including text and artifacts produced with AI assistance.

%
% ---- Bibliography ----
\clearpage
\bibliographystyle{abbrvnat}
\bibliography{main}

\clearpage
\appendix
\section*{Appendix}
\startappendixtoc
\setcounter{tocdepth}{2}
\appendixtableofcontents

% ==============================================================
% APPENDIX
% ==============================================================
\section{Iteration-Matched Controls}
\label[appendix]{app:iteration_matched}

\begin{table}[h!]
  \caption{\textbf{Lower-frequency configurations retain comparable or lower ADE after approximate training-volume matching.} Three-second ADE/FDE are in meters. Each configuration lists sampling frequency $\times$ epochs, followed by ADE/FDE. Training volume is approximated by frequency $\times$ epochs; $\Delta$ is the ADE change relative to the higher-frequency setting, computed before rounding.}
  \label{tab:iteration_matched}
  \centering
  \renewcommand{\arraystretch}{1.08}
  \footnotesize
  \begin{adjustbox}{max width=\linewidth}
    \begin{tabular}{llccc}
    \specialrule{1.2pt}{0pt}{0pt}
    \rowcolor[gray]{.90} 
    \textbf{Model} & \textbf{Dataset} & \textbf{Lower config; ADE/FDE} & \textbf{Higher config; ADE/FDE} & \textbf{$\Delta$ ADE} \\
    \midrule
    E2EDriver & Waymo & 6 $\times$ 8; \textbf{0.518 / 1.473} & 10 $\times$ 5; 0.519 / 1.474 & $-0.12\%$ \\
    E2EDriver & nuScenes & 6 $\times$ 10; \textbf{1.042 / 2.709} & 12 $\times$ 5; 1.185 / 3.096 & $-12.05\%$ \\
    E2EDriver & PAVE & 10 $\times$ 10; \textbf{1.022 / 2.252} & 20 $\times$ 5; 1.083 / 2.374 & $-5.65\%$ \\
    BEV-E2EDriver & Waymo & 6 $\times$ 8; \textbf{0.517 / 1.465} & 10 $\times$ 5; 0.519 / 1.467 & $-0.39\%$ \\
    BEV-E2EDriver & nuScenes & 6 $\times$ 10; \textbf{1.074 / 2.759} & 12 $\times$ 5; 1.189 / 3.098 & $-9.67\%$ \\
    BEV-E2EDriver & PAVE & 10 $\times$ 10; \textbf{1.052 / 2.416} & 20 $\times$ 5; 1.081 / 2.423 & $-2.71\%$ \\
    Tiny-SSR & Waymo & 8 $\times$ 10; \textbf{0.538 / 1.338} & 10 $\times$ 8; 0.559 / 1.399 & $-3.76\%$ \\
    Tiny-SSR & nuScenes & 10 $\times$ 12; \textbf{0.665 / 1.604} & 12 $\times$ 10; 0.726 / 1.706 & $-8.40\%$ \\
    Tiny-SSR & PAVE & 15 $\times$ 16; \textbf{0.528 / 1.106} & 20 $\times$ 12; 0.539 / 1.144 & $-2.04\%$ \\
    \specialrule{1.2pt}{0pt}{0pt}
  \end{tabular}
    \end{adjustbox}
\end{table}

\section{Derivation of the Capacity-Dependent Preferred Temporal Sampling Frequency}
\label[appendix]{app:capacity_derivation}

This section restates the information--capacity model and gives the derivation of the result stated in \Cref{sec:theory}.

\paragraph{Temporal sampling exposes bounded driving-relevant information and increasing driving-irrelevant exposure.}
As described in \Cref{sec:intro}, temporal sampling frequency $f>0$ denotes the rate at which camera frames are sampled from each driving scene. The interval between sampled camera frames is $\Delta t=1/f$. Within each frequency sweep, the scene amount in training set, model architecture, input construction, prediction target, and evaluation protocol remain fixed, while only the temporal sampling frequency varies. This provides a controlled setting for studying how different models utilize increasingly dense temporal exposure.

Let $\mathcal I(f)$ denote the driving-relevant information exposed at frequency $f$. Sparsely sampled camera frames may miss ego motion, agent interactions, and scene evolution, whereas densely sampled camera frames increasingly revisit information already contained within the same driving scenes. Let $\mathcal B(f)$ denote the driving-irrelevant exposure at frequency $f$. For $\Omega>0$, $f_0>0$, $p>1$, and $\nu>0$, we model $\mathcal I(f)$ and $\mathcal B(f)$ as follows:

\begin{equation}
\begin{alignedat}{2}
    \mathcal I(f)
    &=
    \Omega
    \frac{(f/f_0)^p}{1+(f/f_0)^p}
    \qquad&
    \mathcal B(f)
    &\approx
    \nu f.
\end{alignedat}
\label{eq:information-exposure}
\end{equation}

Here, $\Omega$ is the finite limit of driving-relevant information within the fixed driving scenes for the prediction target, $f_0$ is the characteristic temporal sampling frequency governing saturation, and $p$ controls how rapidly $\mathcal I(f)$ changes under sparse temporal sampling. Thus, $\mathcal I(f)\to0$ as $f\to0^+$ and $\mathcal I(f)\to\Omega$ as $f\to\infty$. As temporal sampling becomes denser, sampled camera frames increasingly repeat driving-relevant information already exposed by nearby frames and introduce visual variations unrelated to the driving decision. Following the terminology in \Cref{sec:intro}, we group this redundant content and these unrelated visual variations under \emph{driving-irrelevant exposure}, with $\mathcal B(f)$ growing approximately linearly over the tested range. Thus, driving-relevant information is bounded within the fixed driving scenes, whereas driving-irrelevant exposure can continue to increase with denser temporal sampling. We treat $\mathcal I(f)$ and $\mathcal B(f)$ as phenomenological quantities on a common normalized scale, rather than direct measurements of Shannon information, so that they are comparable and can be combined within the model.

\paragraph{Finite model capacity determines how much driving-relevant information can be utilized.}
Let $\mathcal R(f):=\mathcal I(f)+\mathcal B(f)$ denote the combined modeled exposure at frequency $f$. The fraction of this exposure that is driving-relevant is $\mathcal I(f)/\mathcal R(f)$. We represent finite model capacity by a positive normalized parameter $C$ that is commensurate with $\mathcal R(f)$ within the phenomenological model, and model the fraction of the combined exposure that can be effectively processed as $C/[C+\mathcal R(f)]$. Their product defines a normalized measure of the driving-relevant information effectively utilized by the model:

\begin{equation}
    \alpha_C(f)
    :=
    \widetilde{\mathcal I}_C(f)
    =
    \frac{\mathcal I(f)}{\mathcal R(f)}
    \frac{C}{C+\mathcal R(f)}
    =
    \frac{C\,\mathcal I(f)}
    {\mathcal R(f)\bigl(C+\mathcal R(f)\bigr)}.
    \label{eq:main-effective-information}
\end{equation}

The first factor measures the driving-relevant share of the combined modeled exposure. The second factor specifies how finite model capacity limits the fraction of that exposure that can be effectively processed. Their product is therefore a dimensionless normalized measure of how much of the exposed driving-relevant information can be utilized by a model of capacity $C$ in the presence of driving-irrelevant exposure.

The endpoint limits also ensure that a finite preferred temporal sampling frequency exists. As $f\to0^+$, $\mathcal I(f)\sim\Omega f^p/f_0^p$ and $\mathcal R(f)\sim\nu f$, so

\begin{equation}
\alpha_C(f)\sim\frac{\Omega}{\nu f_0^p}f^{p-1}\longrightarrow0.
\label{eq:appendix-sparse-limit}
\end{equation}

As $f\to\infty$, $\mathcal I(f)\to\Omega$ and $\mathcal R(f)\sim\nu f$, giving

\begin{equation}
\alpha_C(f)\sim\frac{C\Omega}{\nu^2f^2}\longrightarrow0.
\label{eq:appendix-dense-limit}
\end{equation}

Since $\alpha_C(f)>0$ and is continuous for finite $f>0$, it attains a maximum at a finite positive frequency. For the stationary calculation, the information derivative is

\begin{equation}
\mathcal I'(f)
=\frac{\Omega p f_0^p f^{p-1}}{(f_0^p+f^p)^2}>0.
\label{eq:appendix-information-derivative}
\end{equation}

\paragraph{Driving performance increases as more driving-relevant information is effectively utilized.}
Let $\mathcal P_C(f)$ denote a driving-performance metric at frequency $f$. It may be an open-loop trajectory score or a closed-loop driving metric. We orient $\mathcal P_C(f)$ so that larger values mean better performance; an error metric can be negated or transformed monotonically. We assume that, for fixed model capacity, driving performance is monotonically increasing in the normalized measure $\widetilde{\mathcal I}_C(f)$:

\begin{equation}
    \frac{\partial \mathcal P_C(f)}
         {\partial \widetilde{\mathcal I}_C(f)}
    >0.
    \label{eq:main-performance-information}
\end{equation}

Under this monotonicity assumption, maximizing driving performance over temporal sampling frequency is equivalent to maximizing $\widetilde{\mathcal I}_C(f)$:

\begin{equation}
    f_C^\star
    \in
    \underset{f>0}{\arg\max}\;\mathcal P_C(f)
    =
    \underset{f>0}{\arg\max}\;\widetilde{\mathcal I}_C(f).
    \label{eq:main-optimum}
\end{equation}

\paragraph{A capacity-dependent preferred temporal sampling frequency.}
For the derivative, write $I(f):=\mathcal I(f)$ and $R(f):=I(f)+\nu f$, so $R'(f)=I'(f)+\nu$. Applying the quotient rule to Eq.~\eqref{eq:main-effective-information} gives

\begin{equation}
    \frac{\mathrm d\widetilde{\mathcal I}_C(f)}{\mathrm df}
    =
    C\,
    \frac{
        I'(f)R(f)\bigl(C+R(f)\bigr)
        -
        I(f)R'(f)\bigl(C+2R(f)\bigr)
    }
    {\left[R(f)\bigl(C+R(f)\bigr)\right]^2}.
    \label{eq:main-information-derivative}
\end{equation}

At an interior preferred temporal sampling frequency, the derivative vanishes. Hence,

\begin{align}
    0
    &=
    \left.
    \frac{\mathrm d\widetilde{\mathcal I}_C(f)}{\mathrm df}
    \right|_{f=f_C^\star}
    \notag\\
    &\Longleftrightarrow\quad
    I'(f_C^\star)R(f_C^\star)\bigl(C+R(f_C^\star)\bigr)
    \notag\\
    &\qquad =
    I(f_C^\star)R'(f_C^\star)\bigl(C+2R(f_C^\star)\bigr).
    \label{eq:main-stationarity-product}
\end{align}

Dividing by the positive factors yields

\begin{equation}
    \frac{I'(f_C^\star)}{I(f_C^\star)}
    =
    R'(f_C^\star)
    \left[
        \frac{1}{R(f_C^\star)}
        +
        \frac{1}{C+R(f_C^\star)}
    \right].
    \label{eq:main-stationary}
\end{equation}

At a stationary point, the relative increase in driving-relevant information balances the combined effect of increasing modeled exposure in the two denominator factors associated with exposure composition and finite model capacity.

To determine how model capacity changes the preferred temporal sampling frequency, define $G(f,C):=\frac{\partial}{\partial f}\log\widetilde{\mathcal I}_C(f)$. Because $R'(f)=I'(f)+\nu>0$, the capacity derivative of $G$ is

\begin{equation}
    \frac{\partial G(f,C)}{\partial C}
    =
    \frac{R'(f)}{\bigl(C+R(f)\bigr)^2}
    >0.
    \label{eq:main-increasing-differences}
\end{equation}

Suppose that $f_C^\star$ is a nondegenerate interior maximum and changes smoothly with $C$. Then $G(f_C^\star,C)=0$ and $\partial G/\partial f<0$. Implicit differentiation gives

\begin{equation}
    \frac{\mathrm d f_C^\star}{\mathrm d C}
    =
    -
    \frac{\partial G/\partial C}{\partial G/\partial f}
    >0.
    \label{eq:main-capacity-monotonicity}
\end{equation}

Holding the fixed driving scenes and the parameters governing driving-relevant information and driving-irrelevant exposure unchanged, increasing model capacity raises the capacity-dependent preferred temporal sampling frequency under these assumptions. On a finite tested frequency set, the preferred temporal sampling frequency can therefore remain unchanged or shift to a higher tested frequency as model capacity increases. This ordering provides a testable signature of the modeled information--capacity interaction.

For completeness, the discrete result does not require differentiating an optimum. For two tested frequencies $f_2>f_1$, the modeled exposure satisfies $R(f_2)>R(f_1)$. Differentiating their log-effective-information difference gives

\begin{equation}
\frac{\mathrm d}{\mathrm dC}
\left[\log\widetilde{\mathcal I}_C(f_2)-\log\widetilde{\mathcal I}_C(f_1)\right]
=\frac{1}{C+R(f_1)}-\frac{1}{C+R(f_2)}>0.
\label{eq:appendix-discrete-capacity}
\end{equation}

Thus, increasing $C$ favors the higher member of every frequency pair. When $\mathcal I$ and $\mathcal B$ remain fixed across the capacity comparison, a maximizer on a finite tested set cannot move to a lower frequency as capacity increases.

\section{Model and Loss Details}
\label[appendix]{app:model_loss_details}

\paragraph{E2EDriver ($\sim12$M parameters).}
E2EDriver~\citep{MuftiE2EDriverRepo} uses the current multi-view camera-frame at anchor $t$, a 4-second ego-state history, and a high-level navigation command.
A per-view CNN extracts image features, token-level and cross-camera fusion modules form a global visual embedding, and an MLP branch encodes the ego-state history and command.
The fused representation is passed to deterministic trajectory, Gaussian-mixture, and MNLL-style probabilistic heads.
Across temporal sampling frequencies, the current-frame visual input, ego-state history, command encoding, and future ego-trajectory target are unchanged.

\paragraph{Within-family E2EDriver capacity configurations.}
We evaluate Tiny E2EDriver, Original E2EDriver, and Large E2EDriver with internal widths 64, 512, and 1,024. All use the same ImageNet-pretrained SqueezeNet 1.1 visual backbone~\citep{Deng2009ImageNet,Iandola2016SqueezeNet}, input construction, training objectives, and training protocol. Only the internal widths and corresponding adapters change. At each temporal sampling frequency, the training data and hyperparameters are identical across configurations. Tiny E2EDriver is distinct from the Tiny-SSR baseline.

\begin{table}[tb]

  \caption{\textbf{E2EDriver within-family capacity configurations.} The visual backbone and the remaining architectural and training settings are held fixed.}
  \label{tab:e2edriver_width_configs}
  \centering
  \renewcommand{\arraystretch}{1.08}
  \small
  \begin{adjustbox}{max width=\linewidth}
    \begin{tabular}{lrrrl}
    \specialrule{1.2pt}{0pt}{0pt}
    \rowcolor[gray]{.90} 
    \textbf{Model} & \textbf{Internal width} & \textbf{Total parameters} & \textbf{\shortstack[r]{Trainable\\parameters}} & \textbf{\shortstack[l]{Backbone output\\adapter}} \\
    \midrule
    Tiny E2EDriver & 64 & 971,732 & 971,732 & Linear (512$\rightarrow$64) \\
    Original E2EDriver & 512 & 12,203,888 & 12,203,888 & Identity \\
    Large E2EDriver & 1,024 & 46,504,592 & 46,504,592 & Linear (512$\rightarrow$1,024) \\
    \specialrule{1.2pt}{0pt}{0pt}
  \end{tabular}
    \end{adjustbox}
\end{table}

\begin{table}[tb]

  \caption{\textbf{Complete E2EDriver within-family frequency sweeps on NAVTEST.} Results are mean $\pm$ sample standard deviation over three random seeds. Bold identifies the lowest mean ADE/FDE or highest mean EPDMS within each configuration.}
  \label{tab:e2edriver_width_sweep}
  \centering
  \renewcommand{\arraystretch}{1.08}
  \footnotesize
  \begin{adjustbox}{max width=\linewidth}
    \begin{tabular}{llccc}
    \specialrule{1.2pt}{0pt}{0pt}
    \rowcolor[gray]{.90} 
    \textbf{Model} & \textbf{Training frequency} & \textbf{3-s ADE (m) $\downarrow$} & \textbf{3-s FDE (m) $\downarrow$} & \textbf{EPDMS (\%) $\uparrow$} \\
    \midrule
    Tiny E2EDriver & 2\,Hz & 0.5281$\pm$0.0139 & 1.1749$\pm$0.0037 & 69.0919$\pm$0.1539 \\
    Tiny E2EDriver & 4\,Hz & 0.4674$\pm$0.0156 & 1.0369$\pm$0.0037 & 73.4165$\pm$0.1795 \\
    Tiny E2EDriver & 6\,Hz & \textbf{0.4434}$\pm$0.0022 & \textbf{0.9847}$\pm$0.0013 & 75.8087$\pm$0.0989 \\
    Tiny E2EDriver & 8\,Hz & 0.4529$\pm$0.0008 & 0.9975$\pm$0.0021 & 75.0123$\pm$0.1051 \\
    Tiny E2EDriver & 10\,Hz & 0.4454$\pm$0.0010 & 0.9893$\pm$0.0016 & \textbf{77.7761}$\pm$0.1348 \\
    \midrule
    Original E2EDriver & 2\,Hz & 0.395$\pm$0.014 & 0.919$\pm$0.013 & 76.888$\pm$0.298 \\
    Original E2EDriver & 4\,Hz & 0.382$\pm$0.006 & 0.893$\pm$0.005 & 79.138$\pm$0.308 \\
    Original E2EDriver & 6\,Hz & 0.391$\pm$0.024 & 0.898$\pm$0.030 & 80.079$\pm$0.570 \\
    Original E2EDriver & 8\,Hz & \textbf{0.371}$\pm$0.001 & \textbf{0.876}$\pm$0.006 & \textbf{81.558}$\pm$0.078 \\
    Original E2EDriver & 10\,Hz & 0.373$\pm$0.008 & 0.879$\pm$0.007 & 81.343$\pm$0.059 \\
    \midrule
    Large E2EDriver & 2\,Hz & 0.4533$\pm$0.0000 & 1.1282$\pm$0.0005 & 66.5062$\pm$0.0445 \\
    Large E2EDriver & 4\,Hz & 0.4404$\pm$0.0002 & 1.0967$\pm$0.0033 & 67.8559$\pm$0.0049 \\
    Large E2EDriver & 6\,Hz & 0.4320$\pm$0.0008 & 1.0764$\pm$0.0028 & 69.0045$\pm$0.0989 \\
    Large E2EDriver & 8\,Hz & 0.4258$\pm$0.0019 & 1.0575$\pm$0.0044 & 69.6615$\pm$0.1948 \\
    Large E2EDriver & 10\,Hz & \textbf{0.4223}$\pm$0.0005 & \textbf{1.0476}$\pm$0.0012 & \textbf{70.0125}$\pm$0.1161 \\
    \specialrule{1.2pt}{0pt}{0pt}
  \end{tabular}
    \end{adjustbox}
\end{table}

\paragraph{BEV-E2EDriver ($\sim10$M parameters).}
BEV-E2EDriver keeps the ego-state branch, command input, output heads, loss formulation, and future ego-trajectory target of E2EDriver, but replaces the image-space encoder with a BEVFormer-style bird's-eye-view module~\citep{Li2022BEVFormer}.
Multi-view image features from the current anchor are aggregated with BEV queries and spatial cross-attention to produce a global BEV embedding.
This model provides a controlled comparison between image-space and BEV-space perception under the same temporal-frequency intervention.

\paragraph{Tiny-SSR ($\sim18$M parameters).}
Tiny-SSR is a compact variant of the Sparse Scene Representation framework~\citep{Li2025SSR}.
It replaces the baseline ResNet-50 backbone with a ResNet-18~\citep{he2016deep} plus FPN~\citep{lin2017feature}, uses a 256-dimensional embedding, reduces the BEV grid from $100{\times}100$ to $50{\times}50$, and uses one encoder/decoder layer.
Each sample uses the current multi-view frame, one previous visual/BEV context frame at a fixed 0.5-second offset, ego-state information, and a navigation command.
The model predicts a command-conditioned sequence of future ego waypoints over a 3-second horizon at 0.5-second spacing.
Across temporal sampling frequencies, this current-plus-previous-frame input pattern and target horizon are unchanged.

\begingroup
\setlength{\parskip}{3pt}
\titlespacing{\paragraph}{0pt}{4pt}{1em}
\paragraph{AutoVLA ($\sim3$B parameters).}
AutoVLA is instantiated as a Qwen2.5-VL-3B-based multimodal autoregressive trajectory-prediction model~\citep{Bai2025Qwen25VL}.
At each anchor $t$, an AutoVLA sample follows a fixed multimodal template: four temporal image frames ending at $t$, three forward-facing camera views per temporal frame, a fixed system prompt, a fixed multimodal/user prompt, and a fixed ego-state conditioning block.
The future ego-trajectory is represented as action tokens.
A learned action codebook, tokenizer, and rollout decoder convert action tokens back into continuous ego waypoints.
Across the frequency study, the pretrained VLM backbone, visual interface, prompt contract, ego-state pathway, action vocabulary, trajectory decoder, and prediction target remain fixed.
Dataset-specific preprocessing is used only to construct the corresponding frequency-sweep training sets under the original AutoVLA codebase interface.
For the NAVTEST frequency sweeps, we use Qwen2.5-VL-3B-Instruct~\citep{Bai2025Qwen25VL} and the AutoVLA (answer-only SFT) configuration.
\par
\endgroup

\paragraph{Drive-JEPA.}
Drive-JEPA~\citep{wang2026drivejepa} combines a video joint-embedding predictive architecture (V-JEPA) encoder with a proposal-centric trajectory planner. Its published framework pretrains a vision transformer on driving videos, refines multiple waypoint proposals with multimodal trajectory distillation, and selects a trajectory using a momentum-aware score. Within the NAVTEST frequency sweep, the model configuration and per-sample input and prediction target remain fixed. Our NAVTEST configuration is Drive-JEPA (pose-only). We fine-tune the released V-JEPA ViT-L encoder~\citep{Bardes2024VJEPA} with a newly initialized planner and L1 ego-trajectory supervision. We do not use PDM/score or map/collision supervision, because PDM scores are available for the official 2\,Hz data but not for the NAVSIM-10Hz training set. The score-based selection interface remains, but the score branch receives no task-loss supervision.

\paragraph{DriveVLA-W0.}
DriveVLA-W0~\citep{li2025drivevla} augments vision-language-action learning with a world-model objective that predicts future visual content. Its published framework combines a VLA backbone with an action expert for trajectory generation and includes autoregressive visual-token and diffusion-latent world-model variants. Within each frequency comparison, the selected model configuration and trajectory target remain fixed. Our NAVTEST configuration is DriveVLA-W0 (Emu3Pi0). We use NuPlan-pretrained Emu3~\citep{Wang2024Emu3} with the official flow-matching action expert. The VLM backbone is trainable, and downstream training uses the action prediction loss. Visual/world-model supervision comes from pretraining, not from an added future visual prediction loss during these experiments.

\paragraph{HiVT structured-input comparison.}
We evaluate a HiVT-128 adaptation~\citep{Zhou2022HiVT} with 2.56M trainable parameters on a curated subset of the Waymo Open Motion Dataset (WOMD)~\citep{ettinger2021large}. We use complete, approximately 20-second scenarios from the WOMD v1.3.1 \texttt{training\_20s} split. The initial pool of 10,527 scenarios was divided into 8,263 training, 1,061 validation, and 1,203 test scenarios with disjoint scenario IDs. Model-agnostic selection retained 444,643 training anchors from 7,982 training scenarios, matching the NAVSIM 10\,Hz training sample count and approximating its distributions of interval duration, speed, acceleration, turning, and displacement. Within each selected interval, the subset at temporal sampling frequency $f$ retains dense-index positions $\lfloor 10k/f\rfloor$.

Each sample contains 20 observed states at 10\,Hz, including the anchor, and 30 future ego positions over the next three seconds. Inputs include ego and surrounding-agent histories, actor attributes, vectorized lanes, current traffic-light states, and a four-way oracle route command derived from the future ego trajectory. The command is stopped for displacement below 1\,m over three seconds. Otherwise, heading changes above $+15^\circ$ or below $-15^\circ$ indicate left or right, respectively, and the remaining cases are straight. Only the future ego trajectory is supervised. For each of three random seeds, all frequencies start from the same initial weights and are trained from scratch for eight epochs. Each selected anchor is visited once per epoch, including in the final partial batch, so higher frequencies receive more optimization updates. This is a complementary structured-input setting, not a same-model comparison against raw camera inputs.

\begin{table}[tb]

  \caption{\textbf{HiVT structured-input frequency sweep.} All configurations use eight epochs. Three-second trajectory ADE/FDE are in meters, reported as mean $\pm$ reported standard deviation.}
  \label{tab:hivt_structured_sweep}
  \centering
  \renewcommand{\arraystretch}{1.08}
  \begin{adjustbox}{max width=\linewidth}
    \begin{tabular}{rrcc}
    \specialrule{1.2pt}{0pt}{0pt}
    \rowcolor[gray]{.90} 
    \textbf{Training frequency} & \textbf{Epochs} & \textbf{3-s ADE (m) $\downarrow$} & \textbf{3-s FDE (m) $\downarrow$} \\
    \midrule
    2\,Hz & 8 & 0.5847$\pm$0.0019 & 1.6629$\pm$0.0018 \\
    4\,Hz & 8 & 0.5363$\pm$0.0021 & 1.5507$\pm$0.0024 \\
    6\,Hz & 8 & 0.5185$\pm$0.0011 & 1.4914$\pm$0.0019 \\
    8\,Hz & 8 & 0.5305$\pm$0.0008 & 1.5345$\pm$0.0036 \\
    \rowcolor[gray]{.92} 10\,Hz & 8 & \textbf{0.5018}$\pm$0.0014 & \textbf{1.4571}$\pm$0.0021 \\
    \specialrule{1.2pt}{0pt}{0pt}
  \end{tabular}
    \end{adjustbox}
\end{table}

\paragraph{Driving-irrelevant exposure intervention.}
We compare the original training results of Original E2EDriver against two settings with additional driving-irrelevant exposure introduced as noise. Both added-noise settings use Original E2EDriver (width 512; 12,203,888 parameters), the same trainable ImageNet-pretrained SqueezeNet backbone and training objectives, batch size 64, and 10 epochs at each temporal sampling frequency. For image-only noise, we add spatially correlated RGB texture noise at pre-clipping RMS 0.18 before the existing ImageNet normalization, Gaussian noise, and RandomErasing~\citep{Zhong2020RandomErasing} augmentations. The spatial noise is deterministic for each image path across frequencies and epochs. For image-plus-state noise, we additionally add noise to the 16-timestamp ego-state history (position: 1\,m; velocity: 1\,m/s; acceleration: 0.5\,m/s$^2$), including temporally constant, smooth, and independent components. The same noisy history enters the history encoder and all eight camera early-fusion rows. Navigation intent and future ground-truth trajectories are unchanged. All evaluations use clean images and clean history on the same NAVTEST scenarios.

\begin{table}[tb]

  \caption{\textbf{Targeted driving-irrelevant exposure intervention for Original E2EDriver.} Additional exposure is introduced as image-only or image-plus-state noise. Results are mean $\pm$ sample standard deviation over three random seeds. Bold indicates the best mean within each training condition and metric.}
  \label{tab:e2edriver_driving_irrelevant_noise}
  \centering
  \renewcommand{\arraystretch}{1.08}
  \footnotesize
  \begin{adjustbox}{max width=\linewidth}
    \begin{tabular}{llccc}
    \specialrule{1.2pt}{0pt}{0pt}
    \rowcolor[gray]{.90} 
    \textbf{Training condition} & \textbf{Training frequency} & \textbf{EPDMS (\%) $\uparrow$} & \textbf{3-s ADE (m) $\downarrow$} & \textbf{3-s FDE (m) $\downarrow$} \\
    \midrule
    Original training & 4\,Hz & 79.138$\pm$0.308 & 0.382$\pm$0.006 & 0.893$\pm$0.005 \\
    Original training & 6\,Hz & 80.079$\pm$0.570 & 0.391$\pm$0.024 & 0.898$\pm$0.030 \\
    Original training & 8\,Hz & \textbf{81.558}$\pm$0.078 & \textbf{0.371}$\pm$0.001 & \textbf{0.876}$\pm$0.006 \\
    Original training & 10\,Hz & 81.343$\pm$0.059 & 0.373$\pm$0.008 & 0.879$\pm$0.007 \\
    \midrule
    Image-only noise & 4\,Hz & 79.9585$\pm$0.7249 & 0.3743$\pm$0.0029 & 0.8794$\pm$0.0081 \\
    Image-only noise & 6\,Hz & 80.7704$\pm$0.2253 & \textbf{0.3721}$\pm$0.0006 & \textbf{0.8782}$\pm$0.0018 \\
    Image-only noise & 8\,Hz & 80.9530$\pm$0.4114 & 0.3739$\pm$0.0016 & 0.8821$\pm$0.0015 \\
    Image-only noise & 10\,Hz & \textbf{81.3080}$\pm$0.0492 & 0.3722$\pm$0.0021 & 0.8789$\pm$0.0057 \\
    \midrule
    Image + state noise & 4\,Hz & 78.6778$\pm$0.1964 & \textbf{0.6546}$\pm$0.0128 & \textbf{1.3413}$\pm$0.0213 \\
    Image + state noise & 6\,Hz & 78.7034$\pm$0.1069 & 0.6673$\pm$0.0088 & 1.3628$\pm$0.0151 \\
    Image + state noise & 8\,Hz & \textbf{79.2125}$\pm$0.1257 & 0.6924$\pm$0.0066 & 1.4010$\pm$0.0120 \\
    Image + state noise & 10\,Hz & 78.8919$\pm$0.2594 & 0.7219$\pm$0.0006 & 1.4552$\pm$0.0039 \\
    \specialrule{1.2pt}{0pt}{0pt}
  \end{tabular}
    \end{adjustbox}
\end{table}

\paragraph{Training losses.}
For E2EDriver and BEV-E2EDriver, the deterministic trajectory head is supervised with mean-squared error (MSE), the Gaussian-mixture head is supervised with a Gaussian-mixture negative log-likelihood (GMM-NLL), and the modal negative log-likelihood (MNLL) head assigns supervision to the best-matching candidate trajectory mode while learning its mode probability.
When multiple heads are enabled, the objective is
\begin{equation}
    \mathcal{L}_{\text{E2E}}
    =
    \lambda_{\text{mse}}\mathcal{L}_{\text{mse}}
    +
    \lambda_{\text{gmm}}\mathcal{L}_{\text{gmm}}
    +
    \lambda_{\text{mnll}}\mathcal{L}_{\text{mnll}}.
    \label{eq:loss_e2e_app}
\end{equation}
Here, $\mathcal{L}_{\text{E2E}}$ denotes the total training loss for E2EDriver or BEV-E2EDriver, $\mathcal{L}_{\text{mse}}$ denotes the deterministic trajectory MSE loss, $\mathcal{L}_{\text{gmm}}$ denotes the Gaussian-mixture negative log-likelihood loss, and $\mathcal{L}_{\text{mnll}}$ denotes the modal negative log-likelihood loss.
The coefficients $\lambda_{\text{mse}}$, $\lambda_{\text{gmm}}$, and $\lambda_{\text{mnll}}$ are nonnegative loss weights controlling the contribution of the corresponding heads.

Tiny-SSR follows the SSR objective, which combines imitation learning for command-conditioned waypoint prediction with BEV feature reconstruction.
Let $\mathcal{L}_{\text{imi}}$ denote the imitation-learning loss for command-conditioned waypoint prediction.
Let $\hat{\mathbf{B}}_{t+1}$ denote the predicted future BEV feature at the next prediction step, and let $\mathbf{B}_{t+1}$ denote the corresponding target BEV feature:
\begin{equation}
    \mathcal{L}_{\text{bev}}
    =
    \left\lVert
    \hat{\mathbf{B}}_{t+1}
    -
    \mathbf{B}_{t+1}
    \right\rVert_2,
    \qquad
    \mathcal{L}_{\text{SSR}}
    =
    \mathcal{L}_{\text{imi}}
    +
    \mathcal{L}_{\text{bev}}.
    \label{eq:loss_ssr_app}
\end{equation}
Here, $\mathcal{L}_{\text{bev}}$ denotes the BEV feature reconstruction loss, and $\mathcal{L}_{\text{SSR}}$ denotes the total Tiny-SSR training loss.

AutoVLA is trained with supervised fine-tuning (SFT) of the multimodal autoregressive model~\citep{zhou2025autovla}.
Given camera inputs $\mathbf{C}$, prompt tokens $\mathbf{I}$, and ego-state input $\mathbf{S}$, the model predicts the assistant response containing future-trajectory action tokens.
Here, $\mathbf{C}$ denotes the input camera observations, $\mathbf{I}$ denotes the tokenized language prompt, and $\mathbf{S}$ denotes the ego-state input used by the model.
The frequency-response study configuration uses an answer-only target and omits Chain-of-Thought (CoT) reasoning and reinforcement fine-tuning (RFT) from the original AutoVLA methodology:
\begin{equation}
    \mathcal{L}^{\text{SFT}}_i
    =
    \mathcal{L}_{\text{LM},i}
    +
    \lambda_a
    \mathcal{L}_{\text{action},i}.
    \label{eq:loss_autovla_sft_app}
\end{equation}
Here, $i$ indexes a training example, $\mathcal{L}^{\text{SFT}}_i$ denotes the SFT loss for example $i$, $\mathcal{L}_{\text{LM},i}$ denotes the language-modeling loss for the assistant response tokens, and $\mathcal{L}_{\text{action},i}$ denotes the action-token loss.
The coefficient $\lambda_a$ is a nonnegative loss weight for the action-token loss.
Predicted action tokens are decoded into continuous future ego waypoints and evaluated with average displacement error (ADE) and final displacement error (FDE).

In Drive-JEPA's published planner, the training objective combines trajectory-proposal loss with proposal-score, mapping, and collision-prediction losses~\citep{wang2026drivejepa}. Its video encoder is pretrained with the V-JEPA latent-prediction objective. The frequency sweep retains the chosen model and training objectives across settings.

DriveVLA-W0's published world-model variants combine action prediction loss with a visual prediction loss~\citep{li2025drivevla}. The autoregressive variant predicts visual tokens, while the diffusion variant denoises a future-image latent. The chosen objective remains fixed within a frequency comparison.

\paragraph{NAVTEST training epochs and checkpoint selection.}
For the evaluated NAVTEST configurations, we follow the architectures, training hyperparameters, and recommended epoch counts from the original model repositories wherever applicable, making only minimal adaptations. Within each model's frequency sweep, we use the same number of epochs at every temporal sampling frequency and report its final-epoch checkpoint, without tuning or selectively reporting results based on performance. Higher frequencies therefore receive more optimization updates in these main fixed-epoch sweeps. The separate iteration-matched controls compare different epoch counts while matching optimization updates.

\begin{table}[tb]

  \caption{\textbf{Fixed-epoch policy for the NAVTEST frequency sweeps.} All frequencies within each model use the listed epoch count, and the reported checkpoint is the final epoch.}
  \label{tab:navtest_fixed_epochs}
  \centering
  \renewcommand{\arraystretch}{1.08}
  \begin{adjustbox}{max width=\linewidth}
    \begin{tabular}{lrl}
    \specialrule{1.2pt}{0pt}{0pt}
    \rowcolor[gray]{.90} 
    \textbf{Evaluated configuration} & \textbf{Epochs at each frequency} & \textbf{Checkpoint} \\
    \midrule
    Drive-JEPA (pose-only) & 20 & Final epoch \\
    E2EDriver & 10 & Final epoch \\
    AutoVLA (answer-only SFT) & 5 & Final epoch \\
    DriveVLA-W0 (Emu3Pi0) & 4 & Final epoch \\
    \specialrule{1.2pt}{0pt}{0pt}
  \end{tabular}
    \end{adjustbox}
\end{table}

\paragraph{Training hardware.}
We trained on three machines equipped with one H100 GPU, eight A6000 GPUs, and eight H200 GPUs, respectively.

\section{Timestamp-Anchor Construction and Dataset Handling}
\label[appendix]{app:timestamp_anchor}

For each scene timeline, native camera timestamps are sorted in temporal order.
For a requested temporal sampling frequency $f$, the consecutive camera frame timestamp gap is $\Delta t_f=1/f$.
The frequency-induced anchor sequence is obtained by retaining anchors at this interval along the scene timeline, subject to model-specific validity constraints.
The validity operator $\operatorname{Valid}_m(\mathcal{S}_i,t)$ excludes anchors for which the required model input, ego-state history, camera views, previous context frame, or future ego-trajectory supervision is unavailable.
Thus, changing $f$ changes which anchors are retained, but not the semantics of a retained sample.

For irregular timestamps, missing frames, or synchronization failures, the intended behavior is to retain only anchors satisfying the same model-specific input and future-target requirements used at other frequencies.
Exact reconstruction of each frequency-induced training set requires dataset-specific scene identifiers, timestamp manifests, and the validity rules used for every model.

The new NAVSIM comparison starts from the official 2\,Hz training clips~\citep{dauner2024navsim} and uses corresponding nuPlan recordings~\citep{caesar2021nuplan} to construct a denser 10\,Hz training version. The reported 2\,Hz and 10\,Hz results compare models trained with these two anchor densities and evaluated with NAVSIM metrics. This construction makes additional within-scene timestamps available for the frequency intervention.

\paragraph{Adaptive 4\,Hz sampling.}
A4 uses a nominal sampling budget of approximately four anchors per second. About half retain temporal coverage through uniform selection; the remainder favor larger changes in past ego motion. The motion score averages normalized speed change, acceleration, yaw rate, lateral acceleration, jerk, yaw acceleration, curvature, and stop--go transitions over the preceding 1.5 seconds, without future trajectory labels. Motion-selected anchors initially maintain a 0.125-second separation, relaxed only when needed to fill the sampling budget. The achieved temporal sampling frequency is measured from within-scene selected-timestamp gaps, excluding gaps above 2 seconds.

\paragraph{Reproducibility of frequency-specific training sets.}
The supplementary files provide the sampling scripts, including the exact motion-score normalization, selection and tie-breaking rules, and model-specific input and prediction-target validity checks. The final A4 training sets contain 180,304 AutoVLA samples and 214,856 DriveVLA-W0 samples. We have retained scene/log IDs, source image IDs and timestamps, candidate indices, validity records, selected sample lists, and normalization values for exact reconstruction; these records will accompany the training code.

\section{5-Second trajectory-prediction}
\label[appendix]{app:five_second}

The main comparison uses 3-second trajectory-prediction.
We also report 5-second results where measured, because longer horizons introduce additional trajectory uncertainty and may change the preferred temporal sampling frequency.
The 5-second results are broadly consistent with the 3-second frequency-response, although the preferred frequency can shift with prediction horizon.
On Waymo, E2EDriver and BEV-E2EDriver again perform best at 8\,Hz rather than 10\,Hz.
On nuScenes, E2EDriver and BEV-E2EDriver also remain below the highest evaluated frequency, with the best ADE at 6\,Hz and 8\,Hz, respectively.
On PAVE, their best frequencies shift upward, with E2EDriver best at 12\,Hz and BEV-E2EDriver best at 15\,Hz.
AutoVLA performs best at 10\,Hz on Waymo, 12\,Hz on nuScenes, and approximately 18\,Hz on PAVE, where 18\,Hz and 20\,Hz are very close.
Thus, longer-horizon evaluation does not remove the main pattern: the preferred temporal sampling frequency remains model- and dataset-dependent.

\begin{table}[tb]
    \caption{5-second future ego-trajectory prediction (ADE / FDE in meters, single run where available). \textbf{Bold} marks the best ADE setting for each model--dataset pair.}
    \label{tab:future_5s}
    \centering
  \renewcommand{\arraystretch}{1.08}
    \begin{adjustbox}{max width=\linewidth}
    \begin{tabular}{ll|cc|cc|cc}
        \specialrule{1.2pt}{0pt}{0pt}
        \rowcolor[gray]{.90} 
     &  & \multicolumn{2}{c|}{\textbf{Waymo} (10\,Hz)} & \multicolumn{2}{c|}{\textbf{nuScenes} (12\,Hz)} & \multicolumn{2}{c}{\textbf{PAVE} (20\,Hz eval.)} \\
        \cmidrule(lr){3-4} \cmidrule(lr){5-6} \cmidrule(lr){7-8}
        \textbf{Model} & \textbf{Freq.} & \textbf{ADE$\downarrow$} & \textbf{FDE$\downarrow$} & \textbf{ADE$\downarrow$} & \textbf{FDE$\downarrow$} & \textbf{ADE$\downarrow$} & \textbf{FDE$\downarrow$} \\
        \midrule
        \multirow{8}{*}{E2EDriver}
        & 2\,Hz  & 1.731 & 4.605 & 2.492 & 7.715 & --- & --- \\
        & 4\,Hz  & 1.526 & 4.435 & 2.536 & 7.881 & --- & --- \\
        & 6\,Hz  & 1.488 & 4.323 & \textbf{2.327} & \textbf{7.142} & 2.494 & 6.015 \\
        & 8\,Hz  & \textbf{1.471} & \textbf{4.245} & 2.521 & 7.843 & 2.180 & 5.711 \\
        & 10\,Hz & 1.516 & 4.419 & 2.510 & 7.796 & 2.300 & 5.478 \\
        & 12\,Hz & --- & --- & 2.489 & 7.731 & \textbf{1.942} & \textbf{4.916} \\
        & 15\,Hz & --- & --- & --- & --- & 1.982 & 4.942 \\
        & 18--20\,Hz & --- & --- & --- & --- & 2.28--2.48 & 5.44--5.98 \\
        \midrule
        \multirow{8}{*}{\shortstack[l]{BEV-\\E2EDriver}}
        & 2\,Hz  & 1.591 & 4.561 & 2.559 & 7.843 & --- & --- \\
        & 4\,Hz  & 1.565 & 4.516 & 2.541 & 7.807 & --- & --- \\
        & 6\,Hz  & 1.482 & 4.320 & 2.557 & 7.821 & 2.574 & 7.963 \\
        & 8\,Hz  & \textbf{1.464} & \textbf{4.218} & \textbf{2.351} & \textbf{7.197} & 2.311 & 5.758 \\
        & 10\,Hz & 1.486 & 4.320 & 2.506 & 7.762 & 2.330 & 5.759 \\
        & 12\,Hz & --- & --- & 2.514 & 7.748 & 2.254 & 5.700 \\
        & 15\,Hz & --- & --- & --- & --- & \textbf{2.054} & \textbf{5.174} \\
        & 18--20\,Hz & --- & --- & --- & --- & 2.27--2.48 & 5.42--5.69 \\
        \midrule
        \multirow{9}{*}{AutoVLA}
        & 2\,Hz  & 2.831 & 6.975 & 3.236 & 8.621 & 2.482 & 5.665 \\
        & 4\,Hz  & 2.795 & 6.883 & 2.111 & 5.988 & 1.970 & 4.582 \\
        & 6\,Hz  & 2.766 & 6.820 & 1.910 & 5.474 & 1.953 & \textbf{4.353} \\
        & 8\,Hz  & 2.780 & 6.914 & 1.827 & 5.204 & 1.955 & 4.519 \\
        & 10\,Hz & \textbf{2.758} & \textbf{6.833} & 1.811 & 5.173 & 1.929 & 4.436 \\
        & 12\,Hz & --- & --- & \textbf{1.741} & \textbf{4.994} & 1.905 & 4.421 \\
        & 15\,Hz & --- & --- & --- & --- & 1.878 & 4.403 \\
        & 18\,Hz & --- & --- & --- & --- & \textbf{1.878} & 4.377 \\
        & 20\,Hz & --- & --- & --- & --- & 1.880 & 4.485 \\
        \specialrule{1.2pt}{0pt}{0pt}
    \end{tabular}
    \end{adjustbox}
\end{table}

\end{document}